\documentclass[twoside]{article}

\usepackage[accepted]{aistats2026}
\usepackage[round]{natbib} 

\newtheorem{theorem}{Theorem}

\newtheorem{definition}[theorem]{Definition}

\usepackage{mathtools}
\usepackage{amssymb}
\usepackage{amsmath}
\usepackage{booktabs}
\usepackage{multirow}
\usepackage{wrapfig}
\usepackage{authblk}
\usepackage{xcolor}
\usepackage{url}

\usepackage{algorithm} 
\usepackage{algorithmic} 

\begin{document}

\newcommand{\corrauth}{\textsuperscript{$\dagger$}}

%
\runningtitle{RECRAFT: Rethinking Cross-Modal Fine-Tuning}

%
\runningauthor{TK Tran, MC Dao, PL Nguyen, TN Truong, TN Hoang}

\twocolumn[

\aistatstitle{Rethinking Cross-Modal Fine-Tuning: Optimizing the Interaction Between Feature Alignment and Target Fitting}

\vspace{-8mm}
\aistatsauthor{Trong Khiem Tran$^{1,3}$, Manh Cuong Dao$^{2}$, Phi Le Nguyen$^{3}$}

\aistatsauthor{Thao Nguyen Truong$^{4}$, Trong Nghia Hoang$^{1}$\corrauth}\vspace{2mm}

\aistatsaddress{$^{1}$Washington State University, $^{2}$National University of Singapore}\vspace{-8mm}
\aistatsaddress{$^{4}$National Institute of Advanced Industrial Science and Technology}\vspace{-8mm}
\aistatsaddress{$^{3}$Hanoi University of Science and Technology, \corrauth Corresponding author}



]

\begin{abstract}
\vspace{-4mm}
Adapting pre-trained models to unseen feature modalities has become increasingly important due to the growing need for cross-disciplinary knowledge integration.~A key challenge here is how to align the representation of new modalities with the most relevant parts of the pre-trained model's representation space to enable   accurate knowledge transfer.~This requires combining feature alignment with target fine-tuning, but uncalibrated combinations can exacerbate misalignment between the source and target feature-label structures and reduce target generalization.~Existing work however lacks a theoretical understanding of this critical interaction between feature alignment and target fitting.~To bridge this gap, we develop a principled framework that establishes a provable generalization bound on the target error, which explains the interaction between feature alignment and target fitting through a novel concept of feature-label distortion.~This bound offers actionable insights into how this interaction should be optimized for practical algorithm design.~The resulting approach achieves significantly improved performance over state-of-the-art methods across a wide range of benchmark datasets. \vspace{-4mm}
\end{abstract}

\section{Introduction}
\label{sec:intro}\vspace{-2mm}
Modern applications increasingly require transferring representational knowledge from pre-trained foundation models (FMs) to new data modalities. This allows downstream tasks to benefit from the representational knowledge embedded in the pre-trained model when the pre-training data modalities contain relevant information. For example, in genomics, gene expression profiles can be leveraged to enrich the representation for tissue image data~\citep{Lin2024STAlignAM}. Likewise, existing models pre-trained on broad vision~\citep{liu2021swin}, language~\citep{Liu2019RoBERTaAR}, or speech corpora~\citep{radford2022robustspeechrecognitionlargescale} have been increasingly leveraged in domains with new data modalities not seen during pre-training. Recent studies have also shown promising transfers of vision and language models to modalities such as protein structures, cosmic ray signals, and human gestures~\citep{shen2023orca}. 

These early results highlight the potential of cross-modal adaptation and underscore the need for more principled and generalizable approaches. Since a pre-trained FM is optimized to extract the most predictive patterns within its original representation space, positive knowledge transfer requires mapping new data modalities into the most relevant regions of that space. This raises a fundamental challenge:

\noindent {\bf How can we translate new data featuring unseen modalities into an existing pre-trained representation space to enable effective cross-modal knowledge transfer?}

Addressing this challenge is non-trivial, since unlike in-modal fine-tuning, the source and target data distributions often have different statistical structures. Even when the source and target modalities are encoded into the same dimensional space, their resulting feature distributions can differ in covariance structure, higher-order interactions among covariates, and mode geometries. As the pre-trained FM implicitly leverages such distributional information to identify predictive patterns, a mismatch between the source and target representation distributions may cause it to activate spurious or irrelevant patterns, leading to negative transfer. Aligning the representation of new data modalities with relevant distributional geometries of the pre-trained representation space is therefore critical to ensure positive transfer (see Fig.~\ref{feature_select}).

On the other hand, the pre-trained representation space often contains a broad landscape of heterogeneous geometries centered around different modes, many of which may be irrelevant or poorly aligned with the target task. Thus, to avoid negative transfer that harms performance, feature alignment must also be guided by target fitting using fine-tuning data. This guided alignment is non-trivial to formalize, as the interaction between feature alignment and target fitting and its effect on target generalization is not well understood. Existing work addressing this challenge has largely focused on heuristic combinations of feature alignment and target fitting, without providing guarantees on generalization performance for the (downstream) target task~\citep{shen2023orca,cai2024enhancingcrossmodalfinetuninggradually,ma2024learningmodalityknowledgealignment}.\vspace{1mm}

Notably, ORCA~\citep{shen2023orca} aligns source and target representation distributions using optimal transport before fine-tuning the entire network, while PARE~\citep{cai2024enhancingcrossmodalfinetuninggradually} introduces a gating mechanism to combine source and target features during fine-tuning. These methods have demonstrated promising empirical results on benchmarks such as NasBench360~\citep{tu2022nasbench}, but depend on heuristic formulations without explicitly modeling the interaction between feature alignment and target fitting.~MoNA~\citep{ma2024learningmodalityknowledgealignment} characterizes this interaction through a bi-level optimization framework. The inner loop identifies the target-optimal predictor for a given feature embedder, while the outer loop updates the embedder so that its combined representations and predictions align with the source task’s feature-label semantics. This approach aims to reduce the misalignment between the source and target feature-label semantics under a candidate target representation using heuristic alignment measures, while fitting to target data. However, despite its empirical gains, the reliance on heuristic metrics and bi-level design leaves the theoretical link to optimal generalization on the target task unexamined. It remains unclear whether this characterization captures the most effective interaction between feature alignment and target fitting for cross-modal knowledge transfer.\vspace{1mm}

To bridge this gap, we introduce a principled framework that establishes a provable bound on the generalized target error, capturing the interaction between feature alignment and target fitting via a {\bf feature-label distortion} concept. This distortion quantifies the complexity of the probabilistic transport map between the source and target feature-label predictive distributions under a given target feature representation. Intuitively, it measures the cross-modal transferability under a given target representation. A large distortion means low transferability which will cause target fitting to overfit when fine-tuning data is limited, thus decreasing generalized performance.~This provides actionable insights into how this interaction should be optimized, offering a theoretically grounded guide for algorithm design. The above is substantiated with the following technical contributions:\vspace{1mm}

\noindent {\bf Theoretical Analysis.}~We develop a theoretical bound that decomposes the generalized target error into:~(i) the source task error, which serves as a fixed overhead;~(ii) a distributional distance between the source and target representation distributions (i.e., {\bf feature alignment});~(iii) the minimum entropy over a space of probabilistic transport maps between the source and target feature-label conditional distributions (i.e., {\bf feature-label distortion}); and (iv) an alignment term that reflects how well the target predictor follows this transport (i.e., {\bf target fitting}).~To the best of our knowledge, this is the first generalization bound that captures the influence of both the pre-trained model’s quality and the interaction between feature alignment and target fitting on cross-modal fine-tuning performance (Section~\ref{sec:theory}).

\noindent {\bf Algorithm Design.}~We develop a practical algorithm to address the intractability of optimizing the (ii) feature alignment and (iii) feature-label distortion terms over the space of target feature representations and the induced probabilistic transport plans between the source and target feature-label conditional distributions while performing (iv) target fitting.~Our approach constructs an optimizable surrogate that serves as a medium for selectively transferring source knowledge relevant to the target task.~This surrogate is optimized in a preparatory stage to guide the initialization of the main fine-tuning procedure, enabling efficient and targeted cross-modal adaptation (Section~\ref{sec:algo}).

\noindent {\bf Evaluation.}~We evaluate our approach on two comprehensive cross-modal fine-tuning benchmarks:~(1) NAS-Bench-360~\citep{tu2022nasbench}, which covers a broad set of tasks across ten distinct data modalities; and (2) PDEBench~\citep{Takamoto2022PDEBENCHAE}, which assesses model adaptation to simulated data derived from diverse families of partial differential equations (PDEs).~Across both benchmarks, our method consistently outperforms recent state-of-the-art baselines, including ORCA~\citep{shen2023orca}, PARE~\citep{cai2024enhancingcrossmodalfinetuninggradually}, and MoNA~\citep{ma2024learningmodalityknowledgealignment}, on a significant majority of tasks.~These results underscore the importance of designing algorithms guided by an explicit generalization bound framework (Section~\ref{sec:exp}).

\section{Theoretical Analysis}
\label{sec:theory}
This section presents our main theoretical result. We begin by formalizing the cross-modal fine-tuning problem setting and introducing the key notations. We then establish a generalization bound that characterizes how the interaction between feature alignment and target fitting, under a given target feature representation, affects the generalized target performance. This provides a principled foundation for understanding and optimizing cross-modal adaptation (Section~\ref{sec:algo}).

\subsection{Problem Setting and Notations}
\label{sec:notation}
Let $M_s \triangleq (\theta, p_s(z \mid \theta(\boldsymbol{x})))$ denote the learned embedder $\theta$ and the prediction map $p_s(z\mid \theta(\boldsymbol{x}))$ of an FM pre-trained on a source dataset \( (\boldsymbol{X}, \boldsymbol{z}) = \{(\boldsymbol{x}_i, z_i)\}_{i=1}^n \sim D_s(\boldsymbol{x}, z)\).~During fine-tuning, $(\boldsymbol{X}, \boldsymbol{z})$ might not be accessible, but we assume access to an in-modal proxy dataset $(\boldsymbol{X}^{s}, \boldsymbol{z}^s) = \{\boldsymbol{x}^s_i, z_i^s\}_{i=1}^m \sim D_s(\boldsymbol{x}, z)$ which is sampled from the same distribution.

\noindent Let $(\boldsymbol{X}^\tau, \boldsymbol{z}^\tau) = \{\boldsymbol{x}^\tau_i, z_i^\tau\}_{i=1}^\kappa \sim D_\tau(\boldsymbol{x}', z')$ denote the target fine-tuning dataset sampled from another data distribution $D_\tau$ over unseen modalities $\boldsymbol{x}'$ and label $z'$.~We want to construct a target model $M_\tau \triangleq (\phi, p_\tau(z' \mid \phi(\boldsymbol{x}')))$ based on $M_s$ and $(\boldsymbol{X}^\tau, \boldsymbol{z}^\tau) \sim D_\tau(\boldsymbol{x}', z')$ in a principled manner.

\noindent To achieve this, we will establish a mathematical connection (see Theorem~\ref{thm:1}) between the generalized source/target losses (see Definition~\ref{def:1}) and the alignment of their feature distributions (see Definition~\ref{def:2}) under feature maps $\theta$ and $\phi$.

\begin{definition}[Generalized Error]
\label{def:1}
The generalized source/target errors under feature maps $\theta$/$\phi$ are: 
\begin{eqnarray}
\hspace{-10.5mm}\mathrm{err}_s(\theta) &\triangleq& -\mathbb{E}_{(\boldsymbol{x}, z) \sim D_s}\Big[\log p_s\big(z \mid \theta(\boldsymbol{x})\big)\Big] \ ,\\
\hspace{-10mm}\mathrm{err}_\tau(\phi) &\triangleq& -\mathbb{E}_{(\boldsymbol{x}', z') \sim D_\tau}\Big[\log p_\tau\big(z' \mid \phi(\boldsymbol{x}')\big)\Big] \ ,
\end{eqnarray}
which are the expected source and target prediction losses at $(\boldsymbol{x}, z) \sim D_s$ and $(\boldsymbol{x}', z') \sim D_\tau$. 
\end{definition}

\begin{definition}[Feature Distribution]
\label{def:2}
The feature distributions $D_s^\theta(\boldsymbol{u})$/$D_\tau^\phi(\boldsymbol{u})$ are defined as the push-forward of the source's and target's marginal input distributions $D_s(\boldsymbol{x})$/$D_\tau(\boldsymbol{x}')$ under the source and target feature maps, $\boldsymbol{u} = \theta(\boldsymbol{x})$ and $\boldsymbol{u} = \phi(\boldsymbol{x}')$, respectively.
\end{definition}

We also use $D_s^\theta(z\mid \boldsymbol{u})$ and $D_\tau^\phi(z'\mid\boldsymbol{u})$ to denote the source's and target's feature-label conditionals induced from the data distributions $D_s(\boldsymbol{x}, z)$ and $D_\tau(\boldsymbol{x}', z')$ under the source feature map $\boldsymbol{u} = \theta(\boldsymbol{x})$ and the target feature map $\boldsymbol{u} = \phi(\boldsymbol{x}')$, respectively.

\subsection{Main Result}
\label{sec:result}
Our main result characterizes the generalized target loss $\mathrm{err}_\tau(\phi)$ in terms of the following key quantities:\vspace{1mm}

\noindent {\bf Overhead.}~The generalized source loss $\mathrm{err}_s(\theta)$.\vspace{1mm}

\noindent {\bf Feature Alignment (FA).}~A function of the distributional distance between source and target distributions, $D_s^\theta(\boldsymbol{u})$ and $D_\tau^\phi(\boldsymbol{u})$, over feature map $\theta$ and $\phi$. (see Definition~\ref{def:6}).\vspace{1mm} 

\noindent {\bf Feature Label Distortion (FLD).}~A minimum entropy of a valid transport plan $\Lambda^\ast_{\boldsymbol{u}}(z' \mid z)$ over label pairs $(z', z)$ that maps the source conditional $D_s^\theta(z \mid \boldsymbol{u})$ to the target conditional $D_\tau^\phi(z' \mid \boldsymbol{u})$ over the representation $\boldsymbol{u} =  \phi(\boldsymbol{x}')$ produced by the target feature map $\phi$ (see Definition~\ref{def:4}).\vspace{1mm} 

\noindent {\bf Target Fitting (FT).}~A prediction alignment between the target predictor $p_\tau(z' \mid \boldsymbol{u} = \phi(\boldsymbol{x}'))$ and oracle predictor $D_\tau^\phi(z'\mid \boldsymbol{u} = \phi(\boldsymbol{x}'))$ under feature map $\phi$ over the presentation $\boldsymbol{u}= \phi(\boldsymbol{x}')$(see Definition~\ref{def:5}).\vspace{1mm}

\noindent An informal statement of our result is stated below.

\begin{theorem}[Informal Statement]
\label{thm:1}
Under arbitrary feature map $\theta$ and $\phi$, we have:
\begin{eqnarray}
\hspace{-3mm}\mathrm{err}_\tau(\phi) \hspace{-2mm}&\leq&\hspace{-2mm} \mathrm{err}_s(\theta) + \textbf{Feature-Label Distortion}\\
\hspace{-2mm}&+&\hspace{-2mm}{\textbf{Feature Alignment}} + {\textbf{Target Fitting}}. \nonumber 
\end{eqnarray}
\end{theorem}

\noindent This result reveals how the pre-trained model’s quality and the interplay between feature alignment and target fitting shape the cross-modal fine-tuning performance via a feature-label distortion measurement (Definition~\ref{def:4}).~The source loss acts as a fixed overhead, reflecting the influence of the source model.~For FMs, this loss is often negligible.~The remaining terms suggest that aligning features without careful calibration may inadvertently increase the semantic gap between source and target feature-label structures. For instance, when alignment induces representations that enlarge this gap, it can harm generalization by steering target fitting toward overfitting the prediction map to the target data in order to compensate for the poorly aligned representation structure. \vspace{1mm}

\noindent This insight is made precise via the below formal definitions and theorem statement (see Theorem~\ref{thm:2}). 

\begin{definition}[Feature Alignment]
\label{def:6}
Let $\Delta$ denote the set of cost metrics $\delta$ (on the pre-trained representation space) such that the cross-entropy of the source prediction $\ell_s(\boldsymbol{u}) \triangleq -\mathbb{E}_{D^{\theta}_s(z|\boldsymbol{u})}\log p_s(z\mid \boldsymbol{u})$ is $\tau_\delta$-Lipschitz with $\delta$:
\begin{eqnarray}
\hspace{-24mm}\big|\ell_s(\boldsymbol{u}_1) - \ell_s(\boldsymbol{u}_2)\big| &\leq& \tau_\delta \cdot \delta\big(\boldsymbol{u}_1,\boldsymbol{u}_2\big) \ . 
\end{eqnarray}
The feature alignment under target feature map $\phi$ is 
\begin{eqnarray}
\hspace{-9.5mm}\textbf{FA}(\phi,\theta) &\triangleq& \min_{\delta \in \Delta} \Big\{\tau_\delta \cdot W_\delta\Big(D_\tau^\phi(\boldsymbol{u}), D_s^\theta(\boldsymbol{u})\Big) \Big\} \ . \label{eq:FA-def}
\end{eqnarray}
where $W_{\delta}$ is Wasserstein-$1$ distance with cost metric $\delta $ (see definition in Appendix~\ref{sec:wasserstein}).
\end{definition}

\begin{definition}[Feature-Label Distortion]
\label{def:4}
Let $C_{\boldsymbol{u}}^\ast$ denote the set of valid transport plans that satisfy
\begin{eqnarray}
\hspace{-6mm}D_\tau^\phi\big(z'\mid\boldsymbol{u}\big) \hspace{-2mm}&=&\hspace{-2mm} \mathbb{E}_z\Big[\Lambda^\ast_{\boldsymbol{u}}\big(z' \mid z\big)\Big] \ \text{with}\ z \sim D_s^\theta\big(z\mid\boldsymbol{u}\big).\label{eq:FLD-constraint}
\end{eqnarray}
The feature-label distortion at representation $\boldsymbol{u}$ is 
\begin{eqnarray}
\hspace{-18mm}\textbf{FLD}(\boldsymbol{u}) &\triangleq& \min_{\Lambda_{\boldsymbol{u}}^\ast \in C_{\boldsymbol{u}}^\ast}\mathbb{E}_{z}\Big[\mathbb{H}\Big[\Lambda^\ast_{\boldsymbol{u}}\big(z' \mid z\big)\Big] \Big]\ . \label{eq:FLD-def}   
\end{eqnarray}
with $z \sim D_s^\theta\big(z\mid\boldsymbol{u}\big)$.~This characterizes the transferability of source knowledge $\theta$ to target task $\phi$ at $\boldsymbol{u}$.
\end{definition}

\begin{definition}[Target Fitting]
\label{def:5}
Let $C_{\boldsymbol{u}}$ denote the set of valid transport plans $\Lambda_{\boldsymbol{u}}(z' \mid z)$ that satisfy: 
\begin{eqnarray}
\hspace{-7mm}p_\tau\big(z'\mid \boldsymbol{u}\big) \hspace{-2mm}&=&\hspace{-2mm} \mathbb{E}_z\Big[\Lambda_{\boldsymbol{u}}\big(z' \mid z\big)\Big] \ \text{with}\   z \sim D_s^\theta\big(z\mid\boldsymbol{u}\big) .
\end{eqnarray}
The alignment of the target prediction map $p_\tau(z' \mid \boldsymbol{u})$ with the oracle predictor $D_\tau^\phi(z'\mid \boldsymbol{u})$ at $\boldsymbol{u} = \phi(\boldsymbol{x}')$ is
\begin{eqnarray}
\hspace{-0mm}\mathbf{TF}(\boldsymbol{u}) \hspace{-2mm}&\triangleq&\hspace{-2mm} \min_{\Lambda_{\boldsymbol{u}} \in C_{\boldsymbol{u}}} \mathbb{E}_{z}\Big[\mathbb{KL}\Big(\Lambda_{\boldsymbol{u}}^+(.\mid z)\| \Lambda_{\boldsymbol{u}}(.\mid z)\Big) \Big] \ \text{where} \label{eq:TF}
\nonumber\\
\Lambda^+_{\boldsymbol{u}}(.\mid .) \hspace{-2mm}&=&\hspace{-2mm} \arg\hspace{-1mm}\min_{\Lambda^\ast_{\boldsymbol{u}}\in C_{\boldsymbol{u}}^\ast}  \mathbb{E}_{z}\Big[\mathbb{H}\Big[\Lambda^\ast_{\boldsymbol{u}}(z'\mid z)\Big] \Big] .
\end{eqnarray}
\end{definition}
with $z \sim D_s^\theta\big(z\mid\boldsymbol{u}\big)$.~Theorem~\ref{thm:1} is formally stated as:

\begin{theorem}[Formal Statement]
\label{thm:2}
Given the above technical specification of feature alignment, feature-label distortion, and target fitting, we have
\begin{eqnarray}
\hspace{-12mm}\mathrm{err}_\tau(\phi) &\leq&  \mathrm{err}_s(\theta) \ \ + \ \ \textbf{FA}(\phi, \theta)\nonumber\\ 
\hspace{-12mm}&+& \mathbb{E}_{D^{\phi}_{\tau}(\boldsymbol{u})}\Big[\textbf{FLD}(\boldsymbol{u}) \ \ +\ \    \textbf{TF}(\boldsymbol{u}) \Big] \ .\label{eq:bound}
\end{eqnarray}
A detailed proof is provided in Appendix~\ref{app:a}.~Our empirical inspection in Fig.~\ref{fig:loss_components_visualize}, Appendix~\ref{sec:inequality_effective} further shows that the bound is sufficiently tight.
\end{theorem}

Theorem~\ref{thm:2} operationalizes the earlier intuition by providing a complete algorithmic structure to measure the semantic gap in cross-modal fine-tuning under a candidate target representation, captured through both feature alignment and feature-label distortion.~While prior work has relied on distributional alignment to support knowledge transfer into target fitting, the role of feature-label distortion in shaping transferability has been overlooked.~This term exposes the representational discrepancy between the feature-label structures of the source and target domains.~Lower feature-label distortion suggests that the target label can be more readily inferred from the source label information, establishing a consistent pattern that improves transferability from source to target.~This insight reveals that minimizing feature alignment alone may not be sufficient for effective transfer.~We will build on this to design an algorithm that incorporates source-informed regularization to steer away from representations that induce large semantic gap, guiding fine-tuning toward more transferable solutions (Section~\ref{sec:algo}).

\section{Algorithm Design}
\label{sec:algo}
\noindent This section introduces a new cross-modal fine-tuning algorithm named \textbf{RECRAFT} -- \textbf{RE}thinking \textbf{CR}oss-Mod\textbf{A}l \textbf{F}ine-\textbf{T}uning -- which is designed to bridge the semantic gap between source and target tasks in a cross-modal context by optimizing the interaction between feature alignment and target fitting.~It is guided by the result of Theorem~\ref{thm:2}, which bounds the generalized target error by a combination of feature alignment ($\textbf{FA}$) in Eq.~\eqref{eq:FA-def}, target fitting ($\textbf{TF}$) in Eq.~\eqref{eq:TF}, as well as their interaction measured via feature-label distortion ($\textbf{FLD}$) in Eq.~\eqref{eq:FLD-def}.~RECRAFT optimizes this bound via a two-stage workflow as illustrated in Figure~\ref{fig:RECRAFT_overview} below.~Stage $1$ learns the optimal target feature map $\phi$ via minimizing a combination of $\textbf{FA}$ and $\textbf{FLD}$ which defines the source-target semantic gap under $\phi$ (Section~\ref{sec:stage-one}).~Stage $2$ optimizes a target prediction map to minimize the target fitting term $\textbf{TF}$ based on the learned feature map $\phi$ (Section~\ref{sec:stage-two}).

\begin{figure}[htbp]
    \centering
\includegraphics[width=0.48\textwidth]{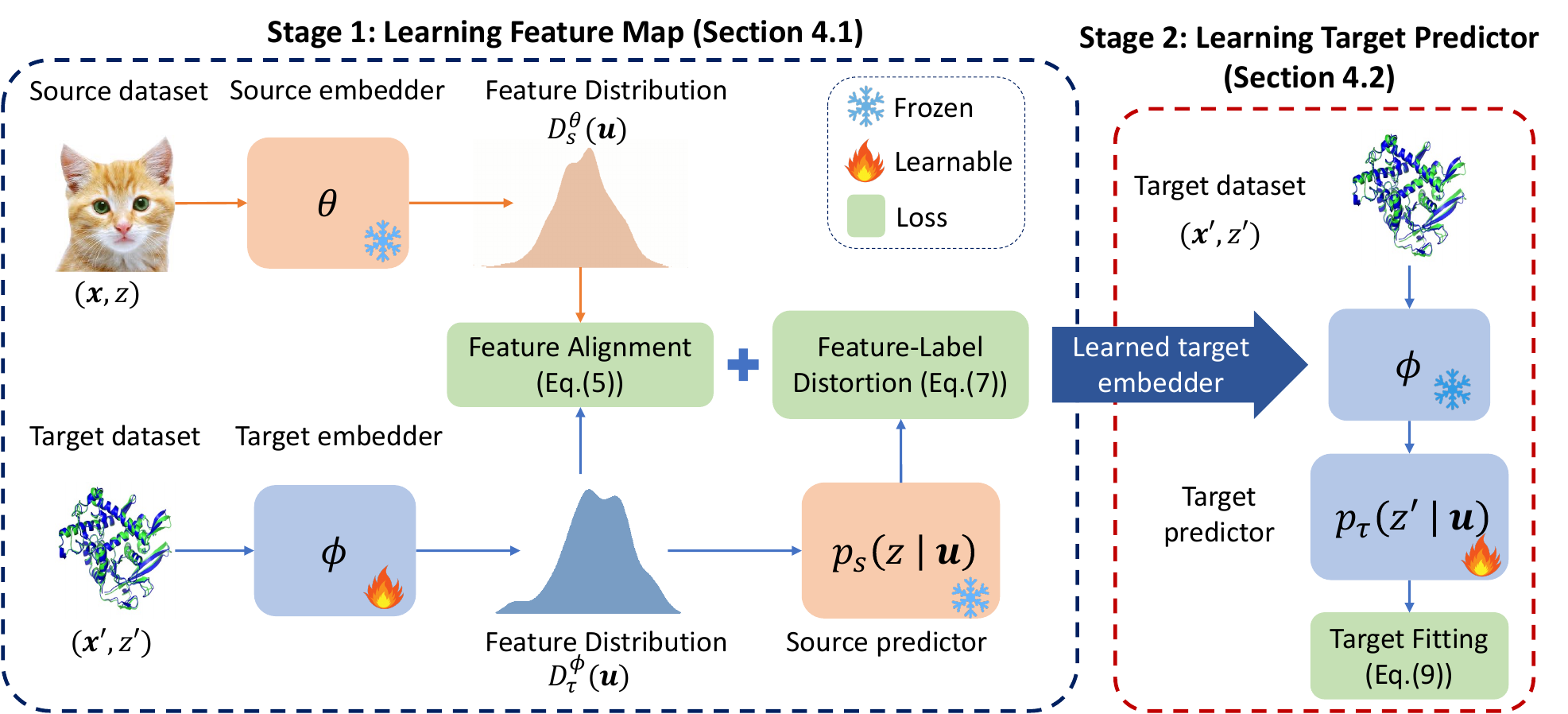}
    \caption{\small Overview of the RECRAFT algorithm.}
    \label{fig:RECRAFT_overview}
\end{figure}

\noindent To elaborate on this design, we note that a direct minimization of the theoretical bound in Eq.~\eqref{eq:bound} of Theorem~\ref{thm:2} is however unstable due to the entangled effect of optimizing both the target prediction map $p_\tau(z' \mid \boldsymbol{u})$ and feature map $\boldsymbol{u} = \phi(\boldsymbol{x}')$ on the oracle and learnable transport sets $C_{\boldsymbol{u}}^\ast$ (see Definition~\ref{def:4}) and $C_{\boldsymbol{u}}$ (see Definition~\ref{def:5}) in complex and interdependent ways.~In particular, changes in the representation $\phi$ simultaneously alter the alignment between source and target feature distributions and reshape the transport landscape $C_{\boldsymbol{u}}^\ast$ on which the target predictor $p_\tau(z'\mid\boldsymbol{u})$ is optimized via minimizing $\textbf{TF}$ in Eq.~\eqref{eq:TF}.

This coupling complicates optimization as it continually moves the optimization target for $p_\tau(z'\mid\boldsymbol{u})$ which destabilizes convergence.~To mitigate this, we adopt a two-stage approach that decomposes the bound minimization into (1) finding a feature map $\phi$ that minimizes the semantic gap $\textbf{FA}(\phi,\theta) + \mathbb{E}_{D^{\phi}_{\tau}(\boldsymbol{u})}[\textbf{FLD}(\boldsymbol{u})]$ between the source and target task, thus maximizing transferability (Section~\ref{sec:stage-one}); and (2) learning a predictor $p_\tau(z'\mid \boldsymbol{u})$ based on the learned $\phi$ (Section~\ref{sec:stage-two}) via minimizing $\mathbb{E}_{D^{\phi}_{\tau}(\boldsymbol{u})}[\textbf{TF}(\boldsymbol{u})]$.~This decomposition stabilizes the optimization by removing interdependencies: the first stage depends only on \( \phi \) while the second stage optimizes \( p_\tau \) given $\phi$, avoiding a moving target.~Figure~\ref{fig:RECRAFT_overview} provides an overview of our algorithm.~Its detailed pseudocode is provided in Appendix~\ref{sec:pseudocode}.

\subsection{Learning Feature Map}
\label{sec:stage-one}
To minimize the semantic gap between source and target, we aim to solve the following:
\begin{eqnarray}
\hspace{-9mm}\phi \hspace{-2mm}&=&\hspace{-2mm} \text{argmin}_\phi \Big\{\textbf{FA}\big(\phi,\theta\big) +  \mathbb{E}_{D^{\phi}_{\tau}(\boldsymbol{u})}\big[\textbf{FLD}(\boldsymbol{u})\big] \Big\} \ .\label{eq:feature}
\end{eqnarray}
As mentioned earlier, this reveals a more principled approach to conditioning the feature map. Prior work such as~\citep{shen2023orca} often minimizes \( \textbf{FA} \) primarily to align the source and target representations, without accounting for its effect on feature-label distortion (\( \textbf{FLD} \)), which captures the semantic misalignment between the source and target feature-label structures induced by \( \phi \).~As \( \textbf{FLD} \) is incorporated into Eq.~\eqref{eq:feature}, our formulation discourages feature maps that increase semantic misalignment which would cause the target fitting stage to compensate improperly during training and reduce transfer performance.~To minimize Eq.~\eqref{eq:feature}, we develop effective optimization surrogates for both $\textbf{FA}$ and $\textbf{FLD}$ as they are not directly tractable.~This is detailed next.

\noindent {\bf A.~Feature Alignment Loss.}~Following Definition~\ref{def:6}, 
\begin{eqnarray}
\hspace{-12mm}\textbf{FA}(\theta,\phi) &=& \min_{\delta \in \Delta} \Big\{\tau_\delta W_\delta\big(D_\tau^\phi(\boldsymbol{u}), D_s^\theta(\boldsymbol{u})\big)\Big\} \ , \label{eq:FA}
\end{eqnarray}
where $\Delta$ is the set of cost metrics $\delta$ for the Wasserstein distance $W_\delta$ such that the cross-entropy source prediction $\ell_s(\boldsymbol{u}) \triangleq -\mathbb{E}_{D_s^\theta(z\mid \boldsymbol{u})}[\log p_s(z\mid \boldsymbol{u})]$ is $\tau_\delta$-Lipschitz with respect to $\delta(\boldsymbol{u}, \boldsymbol{u}')$ (see Definition~\ref{def:6}).\vspace{1mm}

\noindent To effectively constrain the search over $\Delta$ to metric regimes that impose low $\tau_\delta$ on $\ell_s(\boldsymbol{u})$, our main approach is to view $\tau_\delta = \omega$ as a hyperparameter to be determined using a proxy dataset $P_s = (\boldsymbol{X}^s, \boldsymbol{z}^s)$ of the source task.~Given $\omega$, we will adapt the source prediction map $p_s(z\mid\boldsymbol{u}) \simeq p_s(z\mid\boldsymbol{u}; \gamma)$ so that its imposed Lipschitz constant on $\ell_s(\boldsymbol{u})$ under feature map $\boldsymbol{u} = \theta(\boldsymbol{x})$ is around $O(\omega)$.~This is achieved via
\begin{eqnarray}
\hspace{-11.5mm}\text{minimize} \hspace{-2.5mm}&\ &\hspace{-2.5mm} \hspace{-4mm}\underset{\boldsymbol{x}\sim P_s}{\mathbb{E}}\max\Big(0, \big\|\nabla_{\boldsymbol{u}}\ell_s(\theta(\boldsymbol{x}); \gamma)\big\|_\delta - \omega\Big)^2  ,\label{eq:Lipschitz}  
\end{eqnarray}
with respect to $\delta$ and $\omega$.~Here, $\gamma$ can be selected as the parameters of the last layer in the pre-trained prediction map $p_s(z\mid \boldsymbol{u})$.~This generalizes a prior practice on Lipschitz conditioning in~\citep{shen2018wassersteindistanceguidedrepresentation}.~The hyperparameter $\omega$ can be selected to have smallest value without degrading the predictive performance of the source model on the proxy dataset $P_s$.~In our experiment, this optimal value ranges between $0.3$ and $0.5$ across different source models under the Euclidean metric $\delta \equiv \ell_2$ (see Appendix~\ref{sec:FA_assumption_enforce}).~Thus, the feature alignment ($\textbf{FA}$) can be surrogated with:
\begin{eqnarray}
\hspace{-10mm}\textbf{FA}(\theta, \phi) \hspace{-2mm}&\simeq&\hspace{-2mm} L_{\textbf{FA}}(\phi) \triangleq \omega \cdot W_{\ell_2}\Big(D_\tau^\phi(\boldsymbol{u}), D_s^\theta(\boldsymbol{u})\Big) \ ,  \label{eq:FA-loss}
\end{eqnarray}
where $W_{\ell_2}$ is the Wasserstein-$1$ distance with norm-$2$ cost metric $\ell_2$.~See Appendix~\ref{sec:compute_OT} for more details.

\noindent {\bf B.~Feature-Label Distortion Loss.}~We will now construct a surrogate for feature-label distortion, 
\begin{eqnarray}
\hspace{-7mm}\mathbb{E}_{D^{\phi}_{\tau}(\boldsymbol{u})}\Big[\textbf{FLD}(\boldsymbol{u})\Big] \hspace{-3mm}&=&\hspace{-3mm} \mathbb{E}_{(\boldsymbol{u})}\left[\min_{\Lambda_{\boldsymbol{u}}^\ast} \mathbb{E}_{z}\Big[\mathbb{H}\Big[\Lambda_{\boldsymbol{u}}^\ast(. \mid z)\Big]\Big]\right] \\ 
\hspace{-3mm}&\leq&\hspace{-3mm} \mathbb{H}_{(\phi, p_s)}\big[Z' \mid Z,\ \boldsymbol{U}\big]\\
\hspace{-3mm}&\leq&\hspace{-3mm} \mathbb{H}_{(\phi, p_s)}\big[Z' \mid Z\big] \\
\hspace{-3mm}&=&\hspace{-3mm} \mathbb{H}_{(\phi, p_s)}[Z', Z] \ -\ \mathbb{H}_{(\phi, p_s)}[Z]  ,\label{eq:approx}
\end{eqnarray}

where the first inequality hold due to the minimum over the choice of the oracle transport $\Lambda_{\boldsymbol{u}}^\ast(z'\mid z)$ and the definition of conditional entropy.~The second inequality holds due the information-never-hurt property of entropy~\citep{CoverThomas2006}.~The last equality holds from the chain rule.~Eq.~\eqref{eq:approx} allows us to bypass the inaccessible oracle transport $\Lambda_{\boldsymbol{u}}(z'\mid z)$ and estimate it using empirical methods~\citep{nguyen2020leep}.~For each data point $(\boldsymbol{x}', z')$ in the target dataset, we can generate a pseudo source label for it via $z \sim P_s(z \mid \boldsymbol{u})$ under the target feature map $\boldsymbol{u} = \phi(\boldsymbol{x}')$.~Using these statistics, we can approximate $P_{(\phi, p_s)}(z, z') \simeq C(z, z') / \kappa$ where $C(z, z')$ counts the number of times we observe $(z, z')$ via the above pseudo source simulation; and $\kappa$ is the number of target data points.~Likewise, we estimate $P_{(\phi, p_s)}(z) \simeq \sum_{z'} P_{(\phi, p_s)}(z, z')$.~This allows us to approximate
\begin{eqnarray}
\hspace{-0mm}\mathbb{H}_{(\phi, p_s)}[Z', Z] \hspace{-3mm}&=&\hspace{-3mm} -\sum_z\sum_{z'} P_{(\phi, p_s)}\big(z, z'\big)\log P_{(\phi, p_s)}\big(z, z'\big)  \nonumber\\
\mathbb{H}_{(\phi, p_s)}[Z] \hspace{-3mm}&=&\hspace{-3mm} -\sum_z P_{(\phi, p_s)}(z)\log P_{(\phi, p_s)}(z) \ .\label{eq:ent_approx}
\end{eqnarray}
The surrogate for $\textbf{FLD}$ is thus defined as 
\begin{eqnarray}
\hspace{-6mm}\mathbb{E}_{D^{\phi}_{\tau}(\boldsymbol{u})}\Big[\textbf{FLD}(\boldsymbol{u})\Big] \hspace{-2mm}&\simeq&\hspace{-2mm} L_{\textbf{FLD}}(\phi) \triangleq \mathbb{H}_{(\phi, p_s)}\big[Z' \mid Z\big] \\
\hspace{-2mm}&=&\hspace{-2mm} \mathbb{H}_{(\phi, p_s)}[Z', Z] \ -\ \mathbb{H}_{(\phi, p_s)}[Z] .\label{eq:FLD-loss}   
\end{eqnarray}
Combining Eq.~\eqref{eq:FA-loss} and Eq.~\eqref{eq:FLD-loss}, we obtain the following surrogate for Eq.~\eqref{eq:feature},
\begin{eqnarray}
\hspace{-20mm}\phi &=& \text{argmin}_\phi \Big(L_{\textbf{FA}}(\phi) \ +\  L_{\textbf{FLD}}(\phi)\Big)  \ ,\label{eq:feature-loss}  
\end{eqnarray}
which can be minimized effectively using standard numerical optimization method.

\begin{figure*}[t]
    \centering
    \includegraphics[width=0.90\textwidth]{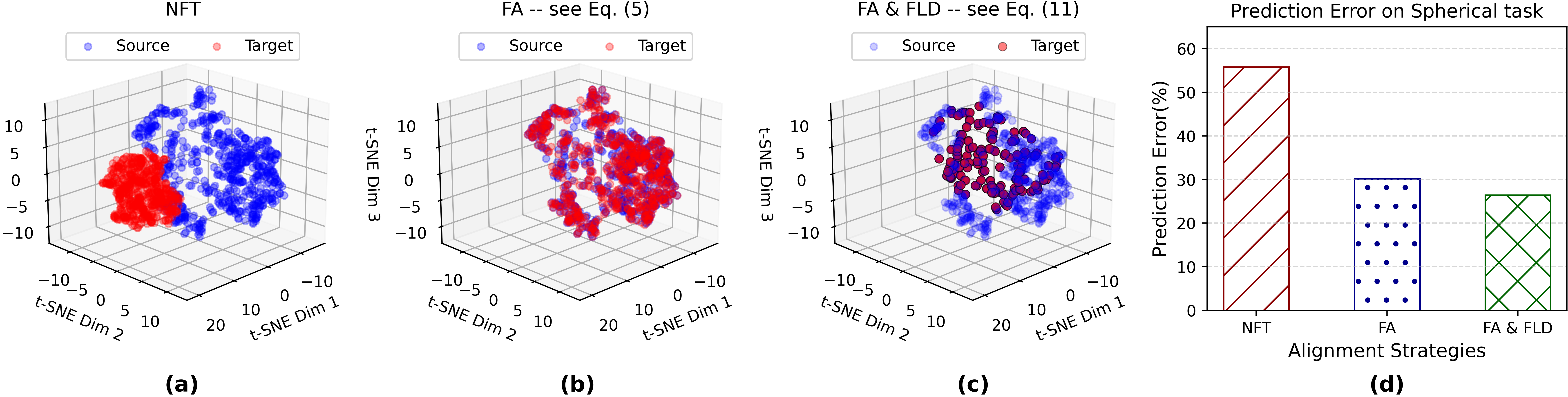} 
    \caption{Visualizations of representation alignment (via tSNE) under $3$ settings: (a) naive fine-tuning (NFT), which ignores alignment; (b) minimization of feature alignment ($\textbf{FA}$) via Eq.~\eqref{eq:FA-def}; and (c) minimizing a sum of $\textbf{FA}$ and feature-label distortion ($\textbf{FLD}$) via Eq.~\eqref{eq:feature}. The corresponding predictive errors are shown in (d).~NFT exhibits no alignment, while minimizing $\textbf{FA}$ leads to exhaustive alignment.~Both results in suboptimal performance.~In contrast, minimizing $\textbf{FA} + \textbf{FLD}$ enables selective alignment and achieves the best performance.
    }
    \label{feature_select}
    \vspace{-4mm}
\end{figure*}

\subsection{Learning Target Predictor}
\label{sec:stage-two}
Given the learned target feature map $\phi$ (Section~\ref{sec:stage-one}), we parameterize the target predictor,
\begin{eqnarray}
\hspace{-7mm}p_\tau(z' \mid \boldsymbol{u}) \hspace{-2mm}&=&\hspace{-2mm} \mathbb{E}_z\big[\Lambda_{\boldsymbol{u}}(z'\mid z)\big] \ \text{with}\ z \sim D_s^\theta(z\mid \boldsymbol{u}) \ ,   
\end{eqnarray}
and a learnable transport $\Lambda_{\boldsymbol{u}}(z'\mid z)$. For convenience, we can also approximate $D_s^\theta(z\mid \boldsymbol{u})$ with $p_s(z\mid \boldsymbol{u})$ since the predictive map $p_s$ of a pre-trained foundation model often capture well the source's feature-label conditional.~Consequently, the learning focuses on $\Lambda_{\boldsymbol{u}}(z'\mid z)$.~We can now parameterize $\Lambda_{\boldsymbol{u}}(z'\mid z) = \Lambda_{\boldsymbol{u}}(z'\mid z; \varphi)$ and optimize its parameterization $\varphi$ via: 
\begin{eqnarray}
\hspace{-7mm}\varphi \hspace{-2mm}&=&\hspace{-2mm} \text{argmin}_\varphi -\mathbb{E}_{(\boldsymbol{x}', z')}\left[ \log p_\tau(z' \mid \phi(\boldsymbol{x}'))  \right] \ \label{target_loss}, \\
\hspace{-7mm}&=&\hspace{-2mm} \text{argmin}_\varphi -\mathbb{E}_{(\boldsymbol{x}', z')}\left[ \log \mathbb{E}_z\Big[\Lambda_{\phi(\boldsymbol{x}')}(z' \mid z; \varphi)\Big]  \right]   
\end{eqnarray}
where $(\boldsymbol{x}', z') \sim (\boldsymbol{X}^\tau, \boldsymbol{z}^\tau)$.~This will make the target predictor $p_\tau(z' \mid \boldsymbol{u}) = \mathbb{E}_z[\Lambda_{\boldsymbol{u}}(z' \mid z; \varphi)]$ approach the target's feature-label conditional $D_\tau^\phi(z'\mid \boldsymbol{u})$. 

\noindent Since $D_\tau^\phi(z'\mid\boldsymbol{u}) = \mathbb{E}_z[\Lambda^\ast_{\boldsymbol{u}}(z' \mid z)]$ (see Definition~\ref{def:4}), aligning $p_\tau(z'\mid \boldsymbol{u})$ with $D_\tau^\phi(z'\mid\boldsymbol{u})$ will decrease the gap between $\Lambda_{\boldsymbol{u}}(z'\mid z; \varphi)$ and $\Lambda_{\boldsymbol{u}}^\ast(z'\mid z)$.~This will in turn reduce the target fitting term (\textbf{TF}) in the target generalization bound in Eq.~\eqref{eq:bound}, Theorem~\ref{thm:2}.~See Appendix~\ref{sec:inequality_effective} for more details.

\section{Empirical Analysis}
\label{sec:exp}
This section presents a detailed empirical evaluation of our proposed method RECRAFT on two popular benchmarks for cross-modal fine-tuning.~These include NASBench-360 \citep{tu2022nasbench} and PDEBench \citep{Takamoto2022PDEBENCHAE}.~NAS-Bench-360 is an extensive benchmark comprising a variety of tasks across 10 distinct data modalities.~PDEBench features simulated data from a variety of partial differential equations (PDEs).~See Appendix~\ref{sec:benmark_info} for additional details.
\subsection{Implementation Details} 
We follow the experiment protocol of ORCA~\citep{shen2023orca}, using  RoBERTa~\citep{Liu2019RoBERTaAR} for 1D tasks and Swin Transformers~\citep{Liu2021SwinTH} for 2D tasks, with CoNLL-2003 and CIFAR-10 as proxy datasets, respectively. Other settings such as learning rates, epochs, and optimizers are taken from ORCA.~Full details are provided in Appendix~\ref{appdendix:hyperparams}.  

\begin{table*}[t]
\caption{Prediction errors (↓) incurred by the tested baselines across tasks in PDEBench~\citep{Takamoto2022PDEBENCHAE}.~The results of MoNA~\citep{ma2024learningmodalityknowledgealignment} are quoted from the corresponding paper due to unavailable code.~RECRAFT achieves best performance on 7 out of 8 tasks and overall average ranking of $1.25$.~The columns  \textbf{\#1} and \textbf{\#2} report the number of times a method achieve best and second best performance, respectively.~See Appendix~\ref{sec:addtional_result} for additional details on the error bars.
}\vspace{2mm}
\label{tab:pde-cross}
\resizebox{0.98\textwidth}{!}{
\begin{tabular}{|l||llllllll||lll|}
\hline
\multirow{3}{*}{\textbf{Model}} 
& \begin{tabular}[l]{@{}l@{}}\textbf{Darcy} \\(2D)\end{tabular} 
& \begin{tabular}[l]{@{}l@{}}\textbf{Advection} \\(1D)\end{tabular} 
& \begin{tabular}[l]{@{}l@{}}\textbf{Burgers} \\(1D)\end{tabular} 
& \begin{tabular}[l]{@{}l@{}}\textbf{Diffusion-}\\ \textbf{Sorption} (1D) \end{tabular} 
& \begin{tabular}[l]{@{}l@{}}\textbf{Shallow }\\ \textbf{Water}  (2D)\end{tabular} 
& \begin{tabular}[l]{@{}l@{}}\textbf{Diffusion-}\\ \textbf{Reaction} (2D) \end{tabular} 
& \begin{tabular}[l]{@{}l@{}}\textbf{Diffusion-}\\ \textbf{Reaction} (1D)  \end{tabular} 
& \begin{tabular}[l]{@{}l@{}}\textbf{Navier-}\\ \textbf{Stokes}  (1D) \end{tabular} 
& \multirow{2}{*}{\begin{tabular}[c]{@{}c@{}}\textbf{Avg.}\\\textbf{Rank}\end{tabular}} 
& \multirow{2}{*}{\textbf{\#1}} 
& \multirow{2}{*}{\textbf{\#2}}\\
& nRMSE & nRMSE& nRMSE& nRMSE& nRMSE& nRMSE& nRMSE& nRMSE\\
\hline
NFT             & $0.085$ & $0.0140$  & $0.0130$  & $3.1\text{\small E-}{3}$ & $6.1\text{\small E-}{3}$ & $0.830$  & $9.2\text{\small E-}{3}$ & $0.863$  & 5.000 & 0 & 0\\
ORCA            & $0.081$ & $0.0098$ & $0.0120$ & $1.8\text{\small E-}{3}$ & $6.0\text{\small E-}{3}$ & $0.820$ & $3.2\text{\small E-}{3}$ & $0.066$  & 3.250 & 0 & 1\\
PARE            & $0.081$ & $\textbf{\textcolor{blue}{0.0032}}$ & $0.0114$ & $1.9\text{\small E-}{3}$ & $5.9\text{\small E-}{3}$ & $0.820$ & $2.9\text{\small E-}{3}$ & $0.068$  & 3.000 & 1 & 5 \\
MoNA & $\textbf{\textcolor{blue}{0.079}}$ & $0.0088$ & $0.0114$ & $\textbf{\textcolor{blue}{1.6\text{\small E-}{3}}}$ & $5.7\text{\small E-}{3}$ & $0.818$ & $\textbf{\textcolor{blue}{2.8\text{\small E-}{3}}}$& $0.054$ & 1.875 & 3 & 4\\
RECRAFT            & $\textbf{\textcolor{blue}{0.079}}$ & $0.0078$ & $\textbf{\textcolor{blue}{0.0108}}$ & $\textbf{\textcolor{blue}{1.6\text{\small E-}{3}}}$ & $\textbf{\textcolor{blue}{5.4\text{\small E-}{3}}}$ & $\textbf{\textcolor{blue}{0.817}}$ & $\textbf{\textcolor{blue}{2.8\text{\small E-}{3}}}$ & $\textbf{\textcolor{blue}{0.050}}$  & \textbf{\textcolor{blue}{1.250}} & \textbf{\textcolor{blue}{7}} & 1\\
\hline
\end{tabular}
}
\end{table*}

\subsection{ Results on NAS-Bench-360} 
The NAS-Bench-360 benchmark~\citep{tu2022nasbench} comprises a diverse set of 10 tasks featuring specialized modalities such as protein sequences, PDE solver, audio, and genetic data, among others.~In this set of experiments, we compare 4 types of baselines: (1) hand-designed solution models~\citep{tu2022nasbench}; (2) general-purpose models that accepted arbitrary inputs converted to byte arrays (without fine-tuning) such as Perceiver IO~\citep{jaegle2021perceiver}; (3) neural architecture search (NAS) methods (no knowledge transfer of existing pre-trained FMs) such as DASH~\citep{shen2022efficientarchitecturesearchdiverse}; and (4) fine-tuning approaches including naive fine-tuning (NFT) and cross-modal fine-tuning methods such as ORCA~\citep{shen2023orca}, PARE~\citep{cai2024enhancingcrossmodalfinetuninggradually}, MoNA~\citep{ma2024learningmodalityknowledgealignment}, and our proposed method RECRAFT.

Table~\ref{tab:nas_results} reports the prediction errors achieved by the above baselines across 10 diverse tasks on the NAS-Bench-360 benchmark.~It can be observed that RECRAFT achieves the lowest prediction error rates on 8 out of 10 tasks, and second lowest error rate on 1 task.~RECRAFT also achieves the best average rank among all baselines.~This demonstrates RECRAFT's robustness in bridging the semantic gap between source and target tasks. We also provide the ablations across several tasks isolating contributions: NFT vs. FA-only vs. RECRAFT, as shown in Table~\ref{tab:complete_nas_FA} (Appendix~\ref{sec:addtional_result}), RECRAFT achieves the best performance across all tasks.~Overall, the result underscores the importance of accounting for feature-label distortion during representation alignment (Definition~\ref{def:4}) which was overlooked in previous work.

\begin{table*}[t]
\setlength{\tabcolsep}{4pt}
\caption{Prediction errors (↓) incurred by the tested baselines across 10 diverse tasks in NAS-Bench-360~\citep{tu2022nasbench}.~The results of MoNA~\citep{ma2024learningmodalityknowledgealignment} are quoted from the corresponding paper due to unavailable code.~RECRAFT achieves best performance on 8 out of 10 tasks which results in the best overall average rank of $1.3$ across all tasks.~The columns \textbf{\#1} and \textbf{\#2} report the number of times a method achieve best and second best performance, respectively. See Appendix~\ref{sec:addtional_result} for additional details on the error bars.
}\vspace{2mm}
\label{tab:nas_results}
\resizebox{1.0\linewidth}{!}{
\begin{tabular}{|l||llllllllll||lll|}
\hline
\multirow{2}{*}{\textbf{Model}} & \textbf{Darcy} & \textbf{DeepSEA} & \textbf{ECG} & \textbf{CIFAR100} & \textbf{Satellite} & \textbf{Spherical} & \textbf{Ninapro} & \textbf{Cosmic} & \textbf{Psicov} & \textbf{FSD50K} &  \multirow{ 2}{*}{\begin{tabular}[c]{@{}c@{}}\textbf{Avg.}\\\textbf{Rank}\end{tabular}} & \multirow{ 2}{*}{\textbf{\#1}} & \multirow{ 2}{*}{\textbf{\#2}}\\
 & Relative $\ell_2$&\text{1-} AUROC   & \text{1-}$\text{F}_1$ score & $\text{0-1 } \text{error } (\%) $ & $\text{0-1 } \text{error } (\%) $ &  $\text{0-1 } \text{error} (\%) $ & $\text{0-1 } \text{error } (\%) $ & \text{1-} AUROC & $\text{MAE}_8$ & \text{1-}mAP & & &\\
\hline
    Hand-designed   & $8.0\text{E-}{3}$ & $0.30$ & $0.28$ & $19.39$ & $19.80$  & $67.41$ & $8.74$ & $0.13$ & $3.37$ & $0.62$ & 5.6 & 0 &1\\
    NAS-Bench-360   & $2.6 \text{E-}{2}$ & $0.32 $ & $0.34$ & $23.39$ & $12.51$ & $48.23$ & $7.35$ & $0.23$ & $2.95$ & $0.63$ & 5.8 & 0 & 0\\
    DASH            & $8.0\text{E-}{3}$ & $0.28$ & $0.32$ & $24.37$ & $12.28$ & $71.38$ & $6.63$ & $0.20$ & $3.30$ & $0.60$ & 4.9  &0 & 2\\
    Perceiver IO    & $2.4 \text{E-}{2}$ & $0.38$ & $0.66$ & $70.04$ & $15.96$ & $82.57$ & $22.4$ & $0.49$ & $8.10$ & $0.73$ & 7.7 & 0 & 0\\
\hline
NFT         & $7.4\text{E-}{3}$ & $0.490$   & $0.44$ & $9.74$  & $13.82$ & $55.76$ & $8.35$ & $0.17$  & $1.92$ & $0.63$ & 5.6 & 0 & 0\\
ORCA       & $7.5\text{E-}{3} $  & $0.291$  & $0.30$ & $7.80$ & $11.63$ & $29.87$ & $7.74$ & $0.15$  & $1.91$ & $0.56$ & 3.8 & 0 & 2\\
PaRE       & $7.4\text{E-}{3}$ & $0.286$  & $0.28$ & $6.70$ & $11.21$ & $27.04$ & $7.12$ & $0.12$  & $\textbf{\textcolor{blue}{0.99}}$ & $\textbf{\textcolor{blue}{0.55}}$ & 2.2 & 2 & 4\\
MoNA & $\textbf{\textcolor{blue}{6.8\text{E-}{3}}}$ & 0.280 & $\textbf{\textcolor{blue}{0.27}}$ & $\textbf{\textcolor{blue}{6.48}}$ & $11.13$ & $27.13$ & $7.28$ & $0.121$ & $\textbf{\textcolor{blue}{0.99}}$ & $\textbf{\textcolor{blue}{0.55}}$ & 1.9 & 5 & 2\\
RECRAFT       & $7.2\text{E-}{3}$& $\textbf{\textcolor{blue}{0.278 }}$  & $\textbf{\textcolor{blue}{0.27}}$ & $7.30$ & $\textbf{\textcolor{blue}{11.11}}$ & $\textbf{\textcolor{blue}{26.41}}$ & $\textbf{\textcolor{blue}{6.60}}$ & $\textbf{\textcolor{blue}{0.11}}$ & $\textbf{\textcolor{blue}{0.99}}$ & $\textbf{\textcolor{blue}{0.55}}$ & \textbf{\textcolor{blue}{1.3}}& \textbf{\textcolor{blue}{8}} & 1\\
\hline
\end{tabular}
}
\end{table*}

\subsection{Results on PDEBench}
PDEBench comprises multiple scientific datasets with simulated data from a wide variety of partial differential equations (PDEs) in physics. We compare the fine-tuning performance of RECRAFT with those of naive fine-tuning (NFT) and prior work on cross-modal fine-tuning methods such as ORCA~\citep{shen2023orca}, PARE~\citep{cai2024enhancingcrossmodalfinetuninggradually}, and MoNA~\citep{ma2024learningmodalityknowledgealignment}.~Table~\ref{tab:pde-cross} reports the prediction error achieved by the above baselines across all tasks.~It is observed that RECRAFT performs best in 7/8 tasks and second best in the remaining task.~It thus achieves the best overall average rank of $1.25$.~This is consistent with our earlier observation on the NAS-Bench-360 benchmark~\citep{tu2022nasbench}, which interestingly demonstrates the effective physic-tuning capability of RECRAFT.~It also outperforms existing specialized physic-informed methods such as Fourier neural operators~\citep{li2021fourier} in 4/8 tasks as shown in Table~\ref{tab:pde-result-expert} in Appendix~\ref{sec:addtional_result}.

\subsection{Impact of Minimizing Semantic Gap}
\begin{figure}[t]
\centering
\includegraphics[width=0.46\textwidth]{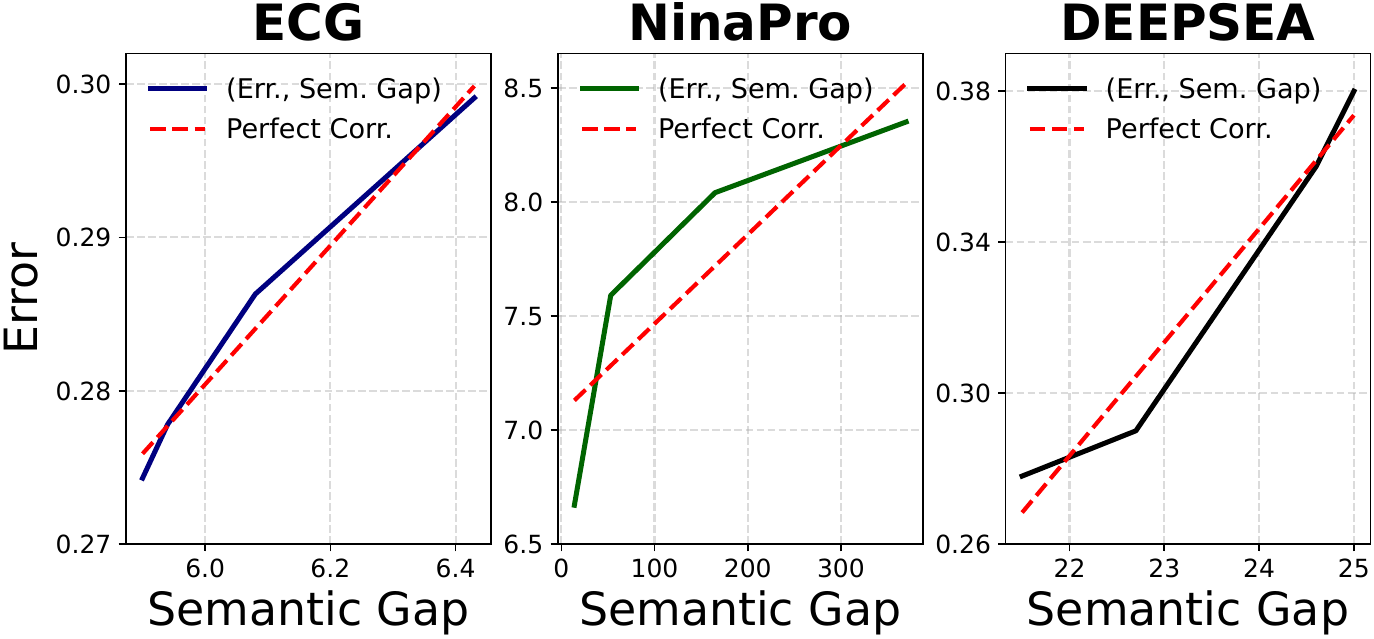}
\caption{Target error versus semantic gap (Eq.~\eqref{eq:feature}) for RECRAFT on ECG, NinaPro, and DeepSEA.~All plots show a strong, consistent correlation across tasks.}\vspace{-4mm}
\label{knowledge_shift}

\end{figure}

This section presents additional empirical evidence to support the insight in Theorem~\ref{thm:2} that minimizing the \textbf{semantic gap} in Eq.~\eqref{eq:feature} (Section~\ref{sec:stage-one}) tightens the upper bound on the generalized target error, thereby improving cross-modal fine-tuning performance.~To validate this connection, we track the prediction error and the corresponding semantic gap across optimization iterations used to learn the feature map in stage 1 (see Section~\ref{sec:stage-one}) on three representative tasks (ECG, NinaPro, and DeepSEA) from NAS-Bench-360. As shown in Fig.~\ref{knowledge_shift}, we observe a strong positive correlation between semantic gap and prediction error, with Pearson correlation coefficients of $0.996$ (ECG), $0.965$ (NinaPro), and $0.989$ (DeepSEA).~These results highlight the practical relevance of the theoretical bound in Theorem~\ref{thm:2} and affirm the value of semantic gap minimization as an effective principle for representation learning in cross-modal fine-tuning.

We also study the effect of incorporating feature-label distortion (see Definition~\ref{def:4}) in the semantic gap formulation of Eq.~\eqref{eq:feature}. This is illustrated in Fig.~\ref{feature_select}, which compares source-target representation alignment under different alignment strategies. Fig.~\ref{feature_select}a and Fig.~\ref{feature_select}b show tSNE feature embeddings under exclusive feature alignment (see Definition~\ref{def:6}) and naive fine-tuning (NFT). NFT produces no alignment with the source's representation space, while exclusive feature alignment leads to an exhaustive alignment.~In contrast, Fig.~\ref{feature_select}c shows that minimizing a combination of feature alignment and feature-label distortion leads to a more selective and effective alignment: target features align only with relevant regions of the source's space. This leads to substantial performance gains over both NFT and exclusive feature alignment (Table~\ref{tab:complete_nas_FA}, Appendix~\ref{sec:addtional_result}).\vspace{-2mm}

\subsection{Systematic Analysis of Existing Methods}
\label{sec:diagnotic_tool}
\begin{figure}[t]
\centering
\includegraphics[width=0.45\textwidth]{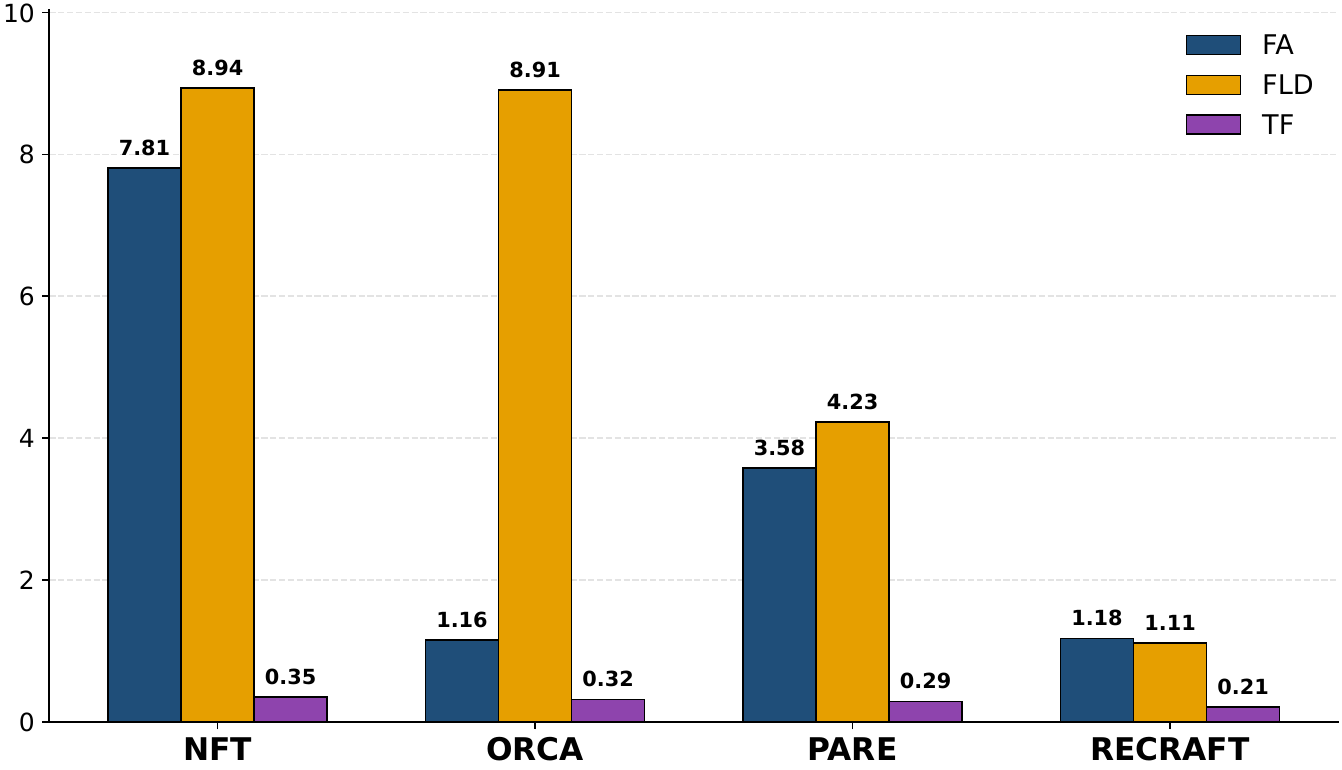}
\caption{Detailed breakdown of the performance bound in Theorem~\ref{thm:2} into {\bf FA}, {\bf FLD}, and {\bf TF} achieved with different cross-modal fine-tuning methods on the Cosmic dataset.~The results show that \textsc{RECRAFT} incurs much lower {\bf FLD} than other baselines.}\vspace{-4mm}


\label{fig:diagnotic_tool}
\end{figure}

The measurable terms {\bf FA} and {\bf FLD} in the theoretical bound on the target generalization error in Theorem~\ref{thm:2} provide a principled diagnostic tool for understanding the strengths and failure modes of prior cross-modal adaptation techniques, which remain largely empirical.~This section provides additional experiments to measure the {\bf FA} (Eq.~\ref{eq:FA-loss}), {\bf FLD} (Eq.~\ref{eq:FLD-def}), and {\bf TF} (Eq.~\ref{eq:TF}) terms incurred by previous methods (Fig.~\ref{fig:diagnotic_tool}).

Based on the results illustrated in Fig.~\ref{fig:diagnotic_tool}, we can draw the following conclusions.~First, NFT (Naive Fine-Tuning) incurs much larger {\bf FA} and {\bf FLD} losses than the other methods and consequently achieves the worst performance (see Tab.~\ref{tab:nas_results} and Tab.~\ref{tab:pde-cross}).~Second, ORCA effectively reduces {\bf FA} but barely reduces {\bf FLD} by an insignificant amount.~This is not surprising given that ORCA performs {\bf FA} and {\bf TF} in separate stages without accounting for their incurred {\bf FLD}.~This can make {\bf TF} overcompensate for a suboptimal {\bf FA}, which increases the risk of overfitting.~Consequently, its effectiveness remains limited compared to our approach. 

Third, PARE has a better balance between {\bf FA} and {\bf FLD} compared to ORCA.~However, its significant reduction on {\bf FLD} appears to come at a cost of increasing {\bf FA} (compared to ORCA), which leads to an overall marginal reduction in {\bf FA} + {\bf FLD}. The significant reduction on {\bf FLD} can be attributed to PARE’s design which aims to minimize a linear combination of source and target losses.~It uses an intermediate distributional representation that combines the most important information from both source and target modalities. Including most important information from the source modalities in the aligned representation helps constrain the feature alignment ({\bf FA}) within the most important regions in the source’s representation landscape, which in turn limits how far the aligned features can deviate from the source’s feature-label structure.~This can help reduce {\bf FLD}.~Nonetheless, this scheme was not optimized to prevent negative transfer, where some source-specific important information is irrelevant to the target’s task but might still be included in the intermediate representation (see Fig.~\ref{feature_select}).~As such, the feature alignment must account for such extra, irrelevant information and hence, increases the {\bf FA} loss.~This explains why PARE incurs larger {\bf FA} than ORCA despite having smaller {\bf FLD}. 

In contrast, our method RECRAFT implements a principled minimization procedure for both {\bf FA + FLD} (Section~\ref{sec:stage-one}) and {\bf TF} (Section~\ref{sec:stage-two}).~This leads to a significant reduction in both {\bf FA} and {\bf FLD} while also reducing {\bf TF}, thus consistently achieving better performance than all the above baselines (Section~\ref{sec:exp}).~Moreover, our theoretical decomposition provides a new analytical lens that will inspire important and synergistic research directions (Appendix~\ref{sec:boarder_impacts}).\vspace{-2mm}

\section{Related Works}\vspace{-2mm}

Due to limited space, we provide a short review of the most relevant work on in-modal and cross-modal fine-tuning. A broader review is deferred to Appendix~\ref{sec:full_related_works}.

\subsection{In-Modal Fine-Tuning of FMs}
\label{sec:in-modal}
The recent advent of large pre-trained or foundation models (FMs) has enabled flexible knowledge transfer to a wide range of downstream tasks under a unified, task-agnostic paradigm. This is commonly known as fine-tuning, which is model-agnostic and does not require shared input or output spaces. Such transfer is feasible because these FMs are trained on broad, diverse corpora (e.g., text, image, or audio) that often overlap semantically or structurally with the content of downstream datasets for in-modal tasks. As a result, fine-tuning has been successfully applied across domains such as vision~\citep{kirillov2023segment, liu2021swin,bui2024revisiting,weng2024probabilistic,Phung2026FederatedPW}, video understanding~\citep{bertasius2021space}, language~\citep{bo2021xtune, yang2022enhancing, ma2023language}, and speech~\citep{radford2022robustspeechrecognitionlargescale, li-etal-2021-multilingual}.~There are also multi-modal FMs~\citep{Radford2021LearningTV,Alayrac2022FlamingoAV,Kim2021ViLTVT} which were pre-trained to learn the embeddings of multiple modalities together but existing approaches to facilitate knowledge transfer from these models are still restricted to within the pre-trained modalities.


\subsection{Cross-Modal Fine-Tuning of FMs}
\label{sec:cross-modal}
Early attempts in cross-modal fine-tuning have focused on transferring language models to other modalities such as vision~\citep{Kiela2019SupervisedMB,tan2019lxmert,gu2022vision}, DNA/protein sequences~\citep{nguyen2023hyenadnalongrangegenomicsequence,lin2023evolutionary,jumper2021highly}.~These provide initial evidence of cross-modal transferability but their designs are hand-tailored to specific target tasks and modalities rather than for general purpose~\citep{shen2023orca}.~Recently, a few general-purpose cross-modal fine-tuning framework have emerged with remarkable successes in finetuning existing vision~\citep{liu2021swin} or language~\citep{Liu2019RoBERTaAR} FMs to solve a broad, diverse set of unseen tasks and data modalities~\citep{shen2023orca,cai2024enhancingcrossmodalfinetuninggradually,ma2024learningmodalityknowledgealignment}.

In particular, ORCA~\citep{shen2023orca} aligns source and target representation distributions using optimal transport before fine-tuning the entire network.~PARE~\citep{cai2024enhancingcrossmodalfinetuninggradually} alternatively introduces a gating mechanism to integrate source and target features during fine-tuning.~MoNA~\citep{ma2024learningmodalityknowledgealignment} addresses modality representation alignment via a bi-level optimization framework: the inner loop learns the optimal target predictor for a given embedder, while the outer loop updates the embedder to align the source feature-label semantics with the combined representation and prediction under this predictor. This approach heuristically reduces the misalignment between source's and target's feature-label structures while fitting to the target data.~However, as noted in Section~\ref{sec:intro}, these approaches largely adopt heuristic methods combining distributional feature alignment and target fitting to facilitate cross-modal knowledge transfer.~Overall, these approaches lack a principled means to assess the impact of their heuristic strategies on the generalized target performance.\vspace{-2mm}

\section{Limitations and Future Works}\vspace{-2mm}
In future work, our proposed method will be further generalized to address two existing limitations.~First, while the use of Euclidean cost metric to characterize the geometric structure under which the Wasserstein distance is computed does not invalidate the bound in Theorem~\ref{thm:2}, it is unclear whether a Wasserstein distance parameterized with Euclidean cost metric would optimally tighten the upper-bound.~We will investigate this aspect via developing a metric learning component which parameterizes and learns this cost metric as additional parameters of the upper-bound.~Second, while upper-bounding {\bf FLD} with the conditional entropic terms in Eq.~\ref{eq:approx} and approximating them with a well-established pseudo-label approach \citep{nguyen2020leep} results in a valid empirical upper-bound, it is unclear whether the bound gap can be made vanishingly small as we increase the no. of sampled pseudo labels (see Eq.~\ref{eq:ent_approx}). This remains an open question to be addressed in the future generalization of our work. \vspace{-4mm}

\section{Conclusion}
\label{sec:conclude}
This paper presents a new theoretical perspective for cross-modal fine-tuning which highlights and addresses a limitation that hinders generalized knowledge transfer in existing work.~This leads to a novel development of a theoretical framework that provably establishes an upper-bound on the generalized target performance.~The bound reveals a precise computational representation for the interaction between feature alignment and target fitting.~This inspires a practical algorithm that optimize this interaction in a principled manner, paving the way for more effective algorithm designs.~The developed algorithm performs significantly better than the recent SOTA methods on two extensive cross-modal fine-tuning benchmark, thus conclusively validating our theoretical insight. 
\newpage

\section*{Acknowledgement}
 This work utilized GPU compute resource at SDSC and ACES through allocation CIS230391 from the Advanced Cyberinfrastructure Coordination Ecosystem: Services and Support (ACCESS) program~\cite{ACCESS-resource}, which is supported by U.S. National Science Foundation grants $\#$2138259, $\#$2138286, $\#$2138307, $\#$2137603, and $\#$2138296.

\bibliographystyle{plainnat} 
\bibliography{reference}

\begin{thebibliography}{49}
\providecommand{\natexlab}[1]{#1}
\providecommand{\url}[1]{\texttt{#1}}
\expandafter\ifx\csname urlstyle\endcsname\relax
  \providecommand{\doi}[1]{doi: #1}\else
  \providecommand{\doi}{doi: \begingroup \urlstyle{rm}\Url}\fi

\bibitem[Alayrac et~al.(2022)Alayrac, Donahue, Luc, Miech, Barr, Hasson, Lenc, Mensch, Millican, Reynolds, Ring, Rutherford, Cabi, Han, Gong, Samangooei, Monteiro, Menick, Borgeaud, Brock, Nematzadeh, Sharifzadeh, Binkowski, Barreira, Vinyals, Zisserman, and Simonyan]{Alayrac2022FlamingoAV}
Jean-Baptiste Alayrac, Jeff Donahue, Pauline Luc, Antoine Miech, Iain Barr, Yana Hasson, Karel Lenc, Arthur Mensch, Katie Millican, Malcolm Reynolds, Roman Ring, Eliza Rutherford, Serkan Cabi, Tengda Han, Zhitao Gong, Sina Samangooei, Marianne Monteiro, Jacob Menick, Sebastian Borgeaud, Andy Brock, Aida Nematzadeh, Sahand Sharifzadeh, Mikolaj Binkowski, Ricardo Barreira, Oriol Vinyals, Andrew Zisserman, and Karen Simonyan.
\newblock Flamingo: a visual language model for few-shot learning.
\newblock \emph{ArXiv}, abs/2204.14198, 2022.
\newblock URL \url{https://api.semanticscholar.org/CorpusID:248476411}.

\bibitem[Bertasius et~al.(2021)Bertasius, Wang, and Torresani]{bertasius2021space}
Gedas Bertasius, Heng Wang, and Lorenzo Torresani.
\newblock Is space-time attention all you need for video understanding?, 2021.

\bibitem[Boerner et~al.(2023)Boerner, Deems, Furlani, Knuth, and Towns]{ACCESS-resource}
Timothy~J. Boerner, Stephen Deems, Thomas~R. Furlani, Shelley~L. Knuth, and John Towns.
\newblock Access: Advancing innovation: Nsf’s advanced cyberinfrastructure coordination ecosystem: Services \& support.
\newblock In \emph{Practice and Experience in Advanced Research Computing 2023: Computing for the Common Good}, PEARC '23, page 173–176, New York, NY, USA, 2023. Association for Computing Machinery.
\newblock ISBN 9781450399852.
\newblock \doi{10.1145/3569951.3597559}.
\newblock URL \url{https://doi.org/10.1145/3569951.3597559}.

\bibitem[Bui et~al.(2024)Bui, Huu, Dinh, Nguyen, and Hoang]{bui2024revisiting}
Long~Minh Bui, Tho~Tran Huu, Duy Dinh, Tan~Minh Nguyen, and Trong~Nghia Hoang.
\newblock Revisiting kernel attention with correlated gaussian process representation.
\newblock In \emph{The 40th Conference on Uncertainty in Artificial Intelligence}, 2024.
\newblock URL \url{https://openreview.net/forum?id=xlIK0vu3MW}.

\bibitem[Cai et~al.(2024)Cai, Li, Ma, Kang, Xie, Sun, and Zhu]{cai2024enhancingcrossmodalfinetuninggradually}
Lincan Cai, Shuang Li, Wenxuan Ma, Jingxuan Kang, Binhui Xie, Zixun Sun, and Chengwei Zhu.
\newblock Enhancing cross-modal fine-tuning with gradually intermediate modality generation, 2024.
\newblock URL \url{https://arxiv.org/abs/2406.09003}.

\bibitem[Cover and Thomas(2006)]{CoverThomas2006}
Thomas~M. Cover and Joy~A. Thomas.
\newblock \emph{Elements of Information Theory}.
\newblock Wiley-Interscience, 2nd edition, July 2006.
\newblock Theorem 2.6.5 states “Conditioning reduces entropy (H(X|Y) \textless{}= H(X)), sometimes summarized as ‘information never hurts.’”.

\bibitem[Gu et~al.(2022)Gu, Chen, Liu, and Li]{gu2022vision}
Junnan Gu, Xi~Chen, Yang Liu, and Ming Li.
\newblock Vision-language pre-training with triple contrastive learning, 2022.

\bibitem[Hoang and Hoang(2024)]{NghiaAAAI24}
Minh Hoang and Trong~Nghia Hoang.
\newblock Few-shot learning via repurposing ensemble of black-box models.
\newblock In \emph{Proc. {AAAI}}, 2024.

\bibitem[Hoang et~al.(2020)Hoang, Lam, Low, and Jaillet]{NghiaICML20}
Trong~Nghia Hoang, Chi~Thanh Lam, Kian~Hsiang Low, and Patrick Jaillet.
\newblock {L}earning {T}ask-{A}gnostic {E}mbedding of {M}ultiple {B}lack-box {E}xperts for {M}ulti-{T}ask {M}odel {F}usion.
\newblock In \emph{Proc. {ICML}}, 2020.

\bibitem[Jaegle et~al.(2021)Jaegle, Borgeaud, Alayrac, Doersch, Ionescu, Ding, Koppula, Zoran, Brock, Shelhamer, et~al.]{jaegle2021perceiver}
Andrew Jaegle, Sebastian Borgeaud, Jean-Baptiste Alayrac, Carl Doersch, Catalin Ionescu, David Ding, Skanda Koppula, Daniel Zoran, Andrew Brock, Evan Shelhamer, et~al.
\newblock Perceiver io: A general architecture for structured inputs \& outputs.
\newblock \emph{arXiv preprint arXiv:2107.14795}, 2021.

\bibitem[Jumper et~al.(2021)Jumper, Evans, Pritzel, Green, Figurnov, Ronneberger, Tunyasuvunakool, Bates, Žídek, Potapenko, Bridgland, Meyer, Kohl, Ballard, Cowie, Romera-Paredes, Nikolov, Jain, Adler, Back, Petersen, Reiman, Clancy, Zielinski, Steinegger, Pacholska, Berghammer, Bodenstein, Silver, Vinyals, Senior, Kavukcuoglu, Kohli, and Hassabis]{jumper2021highly}
John Jumper, Richard Evans, Alexander Pritzel, Tim Green, Michael Figurnov, Olaf Ronneberger, Kathryn Tunyasuvunakool, Russ Bates, Augustin Žídek, Anna Potapenko, Alex Bridgland, Clemens Meyer, Simon A~A Kohl, Andrew~J Ballard, Andrew Cowie, Bernardino Romera-Paredes, Stanislav Nikolov, Rishub Jain, Jonas Adler, Trevor Back, Stig Petersen, David Reiman, Ellen Clancy, Michal Zielinski, Martin Steinegger, Michalina Pacholska, Tamas Berghammer, Sebastian Bodenstein, David Silver, Oriol Vinyals, Andrew~W Senior, Koray Kavukcuoglu, Pushmeet Kohli, and Demis Hassabis.
\newblock Highly accurate protein structure prediction with alphafold, 2021.

\bibitem[Kiela et~al.(2019)Kiela, Bhooshan, Firooz, and Testuggine]{Kiela2019SupervisedMB}
Douwe Kiela, Suvrat Bhooshan, Hamed Firooz, and Davide Testuggine.
\newblock Supervised multimodal bitransformers for classifying images and text.
\newblock \emph{ArXiv}, abs/1909.02950, 2019.
\newblock URL \url{https://api.semanticscholar.org/CorpusID:202539204}.

\bibitem[Kim et~al.(2021)Kim, Son, and Kim]{Kim2021ViLTVT}
Wonjae Kim, Bokyung Son, and Ildoo Kim.
\newblock Vilt: Vision-and-language transformer without convolution or region supervision.
\newblock In \emph{International Conference on Machine Learning}, 2021.
\newblock URL \url{https://api.semanticscholar.org/CorpusID:231839613}.

\bibitem[Kingma and Ba(2017)]{kingma2017adammethodstochasticoptimization}
Diederik~P. Kingma and Jimmy Ba.
\newblock Adam: A method for stochastic optimization, 2017.
\newblock URL \url{https://arxiv.org/abs/1412.6980}.

\bibitem[Kirillov et~al.(2023)Kirillov, Mintun, Ravi, Mao, Rolland, Gustafson, Xiao, Whitehead, Berg, Lo, Doll{\'a}r, and Girshick]{kirillov2023segment}
Alexander Kirillov, Eric Mintun, Nikhila Ravi, Hanzi Mao, Chloe Rolland, Laura Gustafson, Tete Xiao, Spencer Whitehead, Alexander~C Berg, Wan-Yen Lo, Piotr Doll{\'a}r, and Ross Girshick.
\newblock Segment anything, 2023.

\bibitem[Lam et~al.(2021)Lam, Hoang, Low, and Jaillet]{NghiaICML21}
Chi~Thanh Lam, Trong~Nghia Hoang, Kian~Hsiang Low, and Patrick Jaillet.
\newblock {M}odel {F}usion for {P}ersonalized {L}earning.
\newblock In \emph{Proc. {ICML}}, 2021.

\bibitem[Li et~al.(2020)Li, Xie, Wu, Zhao, Liu, and Ding]{Li2020SimultaneousSA}
Shuang Li, Binhui Xie, Jiashu Wu, Ying Zhao, Chi~Harold Liu, and Zhengming Ding.
\newblock Simultaneous semantic alignment network for heterogeneous domain adaptation.
\newblock \emph{Proceedings of the 28th ACM International Conference on Multimedia}, 2020.
\newblock URL \url{https://api.semanticscholar.org/CorpusID:220961739}.

\bibitem[Li et~al.(2021{\natexlab{a}})Li, Wang, Tang, Tran, Tang, Pino, Baevski, Conneau, and Auli]{li-etal-2021-multilingual}
Xian Li, Changhan Wang, Yun Tang, Chau Tran, Yuqing Tang, Juan Pino, Alexei Baevski, Alexis Conneau, and Michael Auli.
\newblock Multilingual speech translation from efficient finetuning of pretrained models.
\newblock In Chengqing Zong, Fei Xia, Wenjie Li, and Roberto Navigli, editors, \emph{Proceedings of the 59th Annual Meeting of the Association for Computational Linguistics and the 11th International Joint Conference on Natural Language Processing (Volume 1: Long Papers)}, pages 827--838, Online, August 2021{\natexlab{a}}. Association for Computational Linguistics.
\newblock \doi{10.18653/v1/2021.acl-long.68}.
\newblock URL \url{https://aclanthology.org/2021.acl-long.68/}.

\bibitem[Li et~al.(2021{\natexlab{b}})Li, Kovachki, Azizzadenesheli, liu, Bhattacharya, Stuart, and Anandkumar]{li2021fourier}
Zongyi Li, Nikola~Borislavov Kovachki, Kamyar Azizzadenesheli, Burigede liu, Kaushik Bhattacharya, Andrew Stuart, and Anima Anandkumar.
\newblock Fourier neural operator for parametric partial differential equations.
\newblock In \emph{International Conference on Learning Representations}, 2021{\natexlab{b}}.
\newblock URL \url{https://openreview.net/forum?id=c8P9NQVtmnO}.

\bibitem[Lin et~al.(2024)Lin, Luo, Chen, Zhang, Wang, Yang, Tong, and Yu]{Lin2024STAlignAM}
Yuxiang Lin, Ling Luo, Ying Chen, Xushi Zhang, Zihui Wang, Wenxian Yang, Mengsha Tong, and Rongshan Yu.
\newblock St-align: A multimodal foundation model for image-gene alignment in spatial transcriptomics.
\newblock \emph{ArXiv}, abs/2411.16793, 2024.
\newblock URL \url{https://api.semanticscholar.org/CorpusID:274280682}.

\bibitem[Lin et~al.(2023)Lin, Akin, Rao, Hie, Zhu, Lu, Smetanin, Verkuil, Kabeli, Shmueli, dos Santos~Costa, Fazel-Zarandi, Sercu, Candido, and Rives]{lin2023evolutionary}
Zeming Lin, Halil Akin, Roshan Rao, Brian Hie, Zhongkai Zhu, Wenting Lu, Nikita Smetanin, Robert Verkuil, Ori Kabeli, Yaniv Shmueli, Allan dos Santos~Costa, Maryam Fazel-Zarandi, Tom Sercu, Salvatore Candido, and Alexander Rives.
\newblock Evolutionary-scale prediction of atomic-level protein structure with a language model.
\newblock \emph{Science}, 379\penalty0 (6637):\penalty0 1123--1130, 2023.
\newblock \doi{10.1126/science.ade2574}.
\newblock URL \url{https://www.science.org/doi/abs/10.1126/science.ade2574}.
\newblock Earlier versions as preprint: bioRxiv 2022.07.20.500902.

\bibitem[Liu et~al.(2019)Liu, Ott, Goyal, Du, Joshi, Chen, Levy, Lewis, Zettlemoyer, and Stoyanov]{Liu2019RoBERTaAR}
Yinhan Liu, Myle Ott, Naman Goyal, Jingfei Du, Mandar Joshi, Danqi Chen, Omer Levy, Mike Lewis, Luke Zettlemoyer, and Veselin Stoyanov.
\newblock Roberta: A robustly optimized bert pretraining approach.
\newblock \emph{ArXiv}, abs/1907.11692, 2019.

\bibitem[Liu et~al.(2021{\natexlab{a}})Liu, Lin, Cao, Hu, Wei, Zhang, Lin, and Guo]{Liu2021SwinTH}
Ze~Liu, Yutong Lin, Yue Cao, Han Hu, Yixuan Wei, Zheng Zhang, Stephen Lin, and Baining Guo.
\newblock Swin transformer: Hierarchical vision transformer using shifted windows.
\newblock \emph{2021 IEEE/CVF International Conference on Computer Vision (ICCV)}, pages 9992--10002, 2021{\natexlab{a}}.

\bibitem[Liu et~al.(2021{\natexlab{b}})Liu, Lin, Cao, Hu, Wei, Zhang, Lin, and Guo]{liu2021swin}
Ze~Liu, Yutong Lin, Yue Cao, Han Hu, Yixuan Wei, Zheng Zhang, Stephen Lin, and Baining Guo.
\newblock Swin transformer: Hierarchical vision transformer using shifted windows, 2021{\natexlab{b}}.

\bibitem[Loshchilov et~al.(2017)Loshchilov, Hutter, et~al.]{loshchilov2017fixing}
Ilya Loshchilov, Frank Hutter, et~al.
\newblock Fixing weight decay regularization in adam.
\newblock \emph{arXiv preprint arXiv:1711.05101}, 5\penalty0 (5):\penalty0 5, 2017.

\bibitem[Ma et~al.(2023)Ma, Li, Cai, and Kang]{ma2023language}
Wenxuan Ma, Shuang Li, Lincan Cai, and Jingxuan Kang.
\newblock Language semantic graph guided data-efficient learning.
\newblock In \emph{Thirty-seventh Conference on Neural Information Processing Systems}, 2023.
\newblock URL \url{https://openreview.net/forum?id=tUyW68cRqr}.

\bibitem[Ma et~al.(2024)Ma, Li, Cai, and Kang]{ma2024learningmodalityknowledgealignment}
Wenxuan Ma, Shuang Li, Lincan Cai, and Jingxuan Kang.
\newblock Learning modality knowledge alignment for cross-modality transfer, 2024.
\newblock URL \url{https://arxiv.org/abs/2406.18864}.

\bibitem[Nguyen et~al.(2020)Nguyen, Hassner, Seeger, and Archambeau]{nguyen2020leep}
Cuong~V. Nguyen, Tal Hassner, Matthias Seeger, and Cedric Archambeau.
\newblock Leep: A new measure to evaluate transferability of learned representations.
\newblock In \emph{Proceedings of the 37th International Conference on Machine Learning}, pages 7294--7305. PMLR, 2020.

\bibitem[Nguyen et~al.(2023)Nguyen, Poli, Faizi, Thomas, Birch-Sykes, Wornow, Patel, Rabideau, Massaroli, Bengio, Ermon, Baccus, and Ré]{nguyen2023hyenadnalongrangegenomicsequence}
Eric Nguyen, Michael Poli, Marjan Faizi, Armin Thomas, Callum Birch-Sykes, Michael Wornow, Aman Patel, Clayton Rabideau, Stefano Massaroli, Yoshua Bengio, Stefano Ermon, Stephen~A. Baccus, and Chris Ré.
\newblock Hyenadna: Long-range genomic sequence modeling at single nucleotide resolution, 2023.
\newblock URL \url{https://arxiv.org/abs/2306.15794}.

\bibitem[Pan and Yang(2010)]{pan2009survey}
Sinno~Jialin Pan and Qiang Yang.
\newblock A survey on transfer learning.
\newblock \emph{IEEE Transactions on Knowledge and Data Engineering}, 22\penalty0 (10):\penalty0 1345--1359, 2010.
\newblock \doi{10.1109/TKDE.2009.191}.

\bibitem[Peyré and Cuturi(2020)]{peyré2020computationaloptimaltransport}
Gabriel Peyré and Marco Cuturi.
\newblock Computational optimal transport, 2020.
\newblock URL \url{https://arxiv.org/abs/1803.00567}.

\bibitem[Phung et~al.(2026)Phung, Nguyen, Huynh, Nguyen, Hoang, and Nguyen]{Phung2026FederatedPW}
Thu~Hang Phung, Duong~M. Nguyen, Thanh~Trung Huynh, Quoc Viet~Hung Nguyen, Trong~Nghia Hoang, and Phi~Le Nguyen.
\newblock Federated prompt-tuning with heterogeneous and incomplete multimodal client data.
\newblock \emph{ArXiv}, abs/2602.07081, 2026.
\newblock URL \url{https://api.semanticscholar.org/CorpusID:284585890}.

\bibitem[Radford et~al.(2021)Radford, Kim, Hallacy, Ramesh, Goh, Agarwal, Sastry, Askell, Mishkin, Clark, Krueger, and Sutskever]{Radford2021LearningTV}
Alec Radford, Jong~Wook Kim, Chris Hallacy, Aditya Ramesh, Gabriel Goh, Sandhini Agarwal, Girish Sastry, Amanda Askell, Pamela Mishkin, Jack Clark, Gretchen Krueger, and Ilya Sutskever.
\newblock Learning transferable visual models from natural language supervision.
\newblock In \emph{International Conference on Machine Learning}, 2021.
\newblock URL \url{https://api.semanticscholar.org/CorpusID:231591445}.

\bibitem[Radford et~al.(2022)Radford, Kim, Xu, Brockman, McLeavey, and Sutskever]{radford2022robustspeechrecognitionlargescale}
Alec Radford, Jong~Wook Kim, Tao Xu, Greg Brockman, Christine McLeavey, and Ilya Sutskever.
\newblock Robust speech recognition via large-scale weak supervision, 2022.
\newblock URL \url{https://arxiv.org/abs/2212.04356}.

\bibitem[Raissi et~al.(2019)Raissi, Perdikaris, and Karniadakis]{RAISSI2019686}
M.~Raissi, P.~Perdikaris, and G.E. Karniadakis.
\newblock Physics-informed neural networks: A deep learning framework for solving forward and inverse problems involving nonlinear partial differential equations.
\newblock \emph{Journal of Computational Physics}, 378:\penalty0 686--707, 2019.
\newblock ISSN 0021-9991.
\newblock \doi{https://doi.org/10.1016/j.jcp.2018.10.045}.
\newblock URL \url{https://www.sciencedirect.com/science/article/pii/S0021999118307125}.

\bibitem[Ronneberger et~al.(2015)Ronneberger, Fischer, and Brox]{10.1007/978-3-319-24574-4_28}
Olaf Ronneberger, Philipp Fischer, and Thomas Brox.
\newblock U-net: Convolutional networks for biomedical image segmentation.
\newblock In Nassir Navab, Joachim Hornegger, William~M. Wells, and Alejandro~F. Frangi, editors, \emph{Medical Image Computing and Computer-Assisted Intervention -- MICCAI 2015}, pages 234--241, Cham, 2015. Springer International Publishing.

\bibitem[Shen et~al.(2018)Shen, Qu, Zhang, and Yu]{shen2018wassersteindistanceguidedrepresentation}
Jian Shen, Yanru Qu, Weinan Zhang, and Yong Yu.
\newblock Wasserstein distance guided representation learning for domain adaptation, 2018.
\newblock URL \url{https://arxiv.org/abs/1707.01217}.

\bibitem[Shen et~al.(2022)Shen, Khodak, and Talwalkar]{shen2022efficientarchitecturesearchdiverse}
Junhong Shen, Mikhail Khodak, and Ameet Talwalkar.
\newblock Efficient architecture search for diverse tasks, 2022.
\newblock URL \url{https://arxiv.org/abs/2204.07554}.

\bibitem[Shen et~al.(2023)Shen, Li, Dery, Staten, Khodak, Neubig, and Talwalkar]{shen2023orca}
Junhong Shen, Liam Li, Lucio~M. Dery, Corey Staten, Mikhail Khodak, Graham Neubig, and Ameet Talwalkar.
\newblock Cross-modal fine-tuning: Align then refine.
\newblock In \emph{Proceedings of the International Conference on Machine Learning (ICML)}. ICML, 2023.
\newblock URL \url{https://arxiv.org/abs/2302.05738}.

\bibitem[Takamoto et~al.(2022)Takamoto, Praditia, Leiteritz, MacKinlay, Alesiani, Pfl{\"u}ger, and Niepert]{Takamoto2022PDEBENCHAE}
Makoto Takamoto, Timothy Praditia, Raphael Leiteritz, Dan MacKinlay, Francesco Alesiani, Dirk Pfl{\"u}ger, and Mathias Niepert.
\newblock Pdebench: An extensive benchmark for scientific machine learning.
\newblock In \emph{Advances in Neural Information Processing Systems (NeurIPS) Datasets and Benchmarks Track}, 2022.

\bibitem[Tan and Bansal(2019)]{tan2019lxmert}
Hao Tan and Mohit Bansal.
\newblock Lxmert: Learning cross-modality encoder representations from transformers, 2019.

\bibitem[Tu et~al.(2022)Tu, Roberts, Khodak, Shen, Sala, and Talwalkar]{tu2022nasbench}
Renbo Tu, Nicholas Roberts, Mikhail Khodak, Junhong Shen, Frederic Sala, and Ameet Talwalkar.
\newblock {NAS}-bench-360: Benchmarking neural architecture search on diverse tasks.
\newblock In \emph{Thirty-sixth Conference on Neural Information Processing Systems Datasets and Benchmarks Track}, 2022.
\newblock URL \url{https://openreview.net/forum?id=xUXTbq6gWsB}.

\bibitem[Villani(2008)]{villani2008optimal}
C.~Villani.
\newblock \emph{Optimal Transport: Old and New}.
\newblock Grundlehren der mathematischen Wissenschaften. Springer Berlin Heidelberg, 2008.
\newblock ISBN 9783540710509.
\newblock URL \url{https://books.google.com.vn/books?id=hV8o5R7_5tkC}.

\bibitem[Wang and Deng(2018)]{wang2018deep}
Mei Wang and Weihong Deng.
\newblock Deep visual domain adaptation: A survey.
\newblock \emph{Neurocomputing}, 312:\penalty0 135--153, 2018.
\newblock ISSN 0925-2312.
\newblock \doi{https://doi.org/10.1016/j.neucom.2018.05.083}.
\newblock URL \url{https://www.sciencedirect.com/science/article/pii/S0925231218306684}.

\bibitem[Weng et~al.(2024)Weng, Hoang, Nguyen, Thai, Weng, and Hoang]{weng2024probabilistic}
Pei-Yau Weng, Minh Hoang, Lam~M. Nguyen, My~T. Thai, Tsui-Wei Weng, and Trong~Nghia Hoang.
\newblock Probabilistic federated prompt-tuning with non-{IID} and imbalanced data.
\newblock In \emph{The Thirty-eighth Annual Conference on Neural Information Processing Systems}, 2024.
\newblock URL \url{https://openreview.net/forum?id=nw6ANsC66G}.

\bibitem[Yang et~al.(2022)Yang, Chen, Zhou, and Li]{yang2022enhancing}
Huiyun Yang, Huadong Chen, Hao Zhou, and Lei Li.
\newblock Enhancing cross-lingual transfer by manifold mixup.
\newblock In \emph{International Conference on Learning Representations}, 2022.
\newblock URL \url{https://openreview.net/forum?id=OjPmfr9GkVv}.

\bibitem[Yao et~al.(2019)Yao, Zhang, Li, and Ye]{Yao2019HeterogeneousDA}
Yuan Yao, Yu~Zhang, Xutao Li, and Yunming Ye.
\newblock Heterogeneous domain adaptation via soft transfer network.
\newblock \emph{Proceedings of the 27th ACM International Conference on Multimedia}, 2019.
\newblock URL \url{https://api.semanticscholar.org/CorpusID:201651510}.

\bibitem[Yin et~al.(2024)Yin, Yu, Lin, Liu, Sonke, and Gavves]{yin2024domainadaptationcauchyschwarzdivergence}
Wenzhe Yin, Shujian Yu, Yicong Lin, Jie Liu, Jan-Jakob Sonke, and Efstratios Gavves.
\newblock Domain adaptation with cauchy-schwarz divergence, 2024.
\newblock URL \url{https://arxiv.org/abs/2405.19978}.

\bibitem[Zheng et~al.(2021)Zheng, Dong, Huang, Wang, Chi, Singhal, Che, Liu, Song, and Wei]{bo2021xtune}
Bo~Zheng, Li~Dong, Shaohan Huang, Wenhui Wang, Zewen Chi, Saksham Singhal, Wanxiang Che, Ting Liu, Xia Song, and Furu Wei.
\newblock {Consistency Regularization for Cross-Lingual Fine-Tuning}.
\newblock In \emph{Proceedings of ACL 2021}, 2021.

\end{thebibliography}

\clearpage

\section*{Checklist}

\begin{enumerate}
  \item For all models and algorithms presented, check if you include:
  \begin{enumerate}
    \item A clear description of the mathematical setting, assumptions, algorithm, and/or model. [Yes]
    \item An analysis of the properties and complexity (time, space, sample size) of any algorithm. [Yes]
    \item (Optional) Anonymized source code, with specification of all dependencies, including external libraries. [Yes]
  \end{enumerate}

  \item For any theoretical claim, check if you include:
  \begin{enumerate}
    \item Statements of the full set of assumptions of all theoretical results. [Yes]
    \item Complete proofs of all theoretical results. [Yes]
    \item Clear explanations of any assumptions. [Yes]     
  \end{enumerate}

  \item For all figures and tables that present empirical results, check if you include:
  \begin{enumerate}
    \item The code, data, and instructions needed to reproduce the main experimental results (either in the supplemental material or as a URL). [Yes, see Appendix ~\ref{appdendix:hyperparams}]
    \item All the training details (e.g., data splits, hyperparameters, how they were chosen). [Yes]
    \item A clear definition of the specific measure or statistics and error bars (e.g., with respect to the random seed after running experiments multiple times). [Yes]
    \item A description of the computing infrastructure used. (e.g., type of GPUs, internal cluster, or cloud provider). [ Yes, see Appendix~\ref{appdendix:hyperparams} ]
  \end{enumerate}

  \item If you are using existing assets (e.g., code, data, models) or curating/releasing new assets, check if you include:
  \begin{enumerate}
    \item Citations of the creator If your work uses existing assets. [Yes]
    \item The license information of the assets, if applicable. [Not Applicable]
    \item New assets either in the supplemental material or as a URL, if applicable. [Yes]
    \item Information about consent from data providers/curators. [Not Applicable, Public datasets]
    \item Discussion of sensible content if applicable, e.g., personally identifiable information or offensive content. [Not Applicable]
  \end{enumerate}

  \item If you used crowdsourcing or conducted research with human subjects, check if you include:
  \begin{enumerate}
    \item The full text of instructions given to participants and screenshots. [Not Applicable]
    \item Descriptions of potential participant risks, with links to Institutional Review Board (IRB) approvals if applicable. [Not Applicable]
    \item The estimated hourly wage paid to participants and the total amount spent on participant compensation. [Not Applicable]
  \end{enumerate}

\end{enumerate}

\clearpage
\appendix
\thispagestyle{empty}

\onecolumn
\aistatstitle{Supplementary Materials}

\section{Proof of Theorem~\ref{thm:2}}
\label{app:a}
This section provides the detailed proof of our main result stated in Theorem~\ref{thm:2}, which bounds the gap between the target and source generalized errors. The bound is characterized in terms of (i) the distributional feature alignment ({\bf FA}) between the source and target in Eq.~\eqref{eq:FA-def}, (ii) the feature-label distortion ({\bf FLD}) under their respective feature maps in Eq.~\eqref{eq:FLD-def}, and (iii) the target model’s fit ({\bf TF}) to the fine-tuning data in Eq.~\eqref{eq:TF}. For clarity, we restate the result below.
\begin{eqnarray}
\mathrm{err}_\tau(\phi) &\leq&  \mathrm{err}_s(\theta) \ \ \ +\ \ \ \textbf{FA}(\phi, \theta)\ +\  \mathbb{E}_{D^{\phi}_{\tau}(\boldsymbol{u})}\Big[\textbf{FLD}(\boldsymbol{u}) \ \ \ +\ \ \    \textbf{TF}(\boldsymbol{u}) \Big] \ ,\label{eq:a1-1}
\end{eqnarray}
Here, $D_\tau^\phi(\boldsymbol{u})$ denote the target's marginal feature distribution under (target) feature map $\phi$ and $D_s^\theta(z\mid \boldsymbol{u})$ denote the feature-label conditional under (source) feature map $\theta$.~Our proof goes below.\vspace{1mm}

\noindent First, following Definition~\ref{def:1}, the generalized target error is
\begin{eqnarray}   \mathrm{err}_\tau(\phi) &\triangleq& -\mathbb{E}_{(\boldsymbol{x}', z') \sim D_\tau}\Big[\log p_\tau\big(z' \mid \phi(\boldsymbol{x}')\big)\Big] \\
&=& -\mathbb{E}_{(\boldsymbol{u}, z') \sim D^{\phi}_\tau}\Big[\log p_\tau\big(z' \mid \boldsymbol{u}\big)\Big]\ =\ -\mathbb{E}_{D^{\phi}_\tau(\boldsymbol{u})}\mathbb{E}_{D^{\phi}_\tau(z' \mid \boldsymbol{u})}\Big[\log p_\tau\big(z' \mid \boldsymbol{u}\big)\Big] \ .\label{eq:a1-2}
\end{eqnarray}
Likewise, the generalized source error is
\begin{eqnarray}   \hspace{-13mm}\mathrm{err}_s(\theta) &=& -\mathbb{E}_{D^{\theta}_s(\boldsymbol{u})}\mathbb{E}_{D^{\theta}_s(z \mid \boldsymbol{u})}\Big[\log p_s\big(z \mid \boldsymbol{u}\big)\Big] \ .\label{eq:a1-3}
\end{eqnarray}
Combining Eqs.~\eqref{eq:a1-2} and~\eqref{eq:a1-3}, we can rewrite
\begin{eqnarray}
\hspace{-28.5mm}\mathrm{err}_\tau(\phi) - \mathrm{err}_s(\theta) &=& \mathbf{A} + \mathbf{B} \quad \text{where} \label{eq:a1-4}
\end{eqnarray}
\begin{eqnarray}
\hspace{-15mm}\mathbf{A} 
\hspace{-2mm}&\triangleq&\hspace{-2mm} \mathrm{err}_\tau(\phi) +  \mathbb{E}_{ D^{\phi}_\tau(\boldsymbol{u})}\mathbb{E}_{ D^{\theta}_s(z\mid \boldsymbol{u})}\log D^{\theta}_s(z \mid \boldsymbol{u})  \ ,\label{eq:a1-5} \\ 
\hspace{-15mm}\mathbf{B}\hspace{-2mm}&\triangleq&\hspace{-2mm} -\mathrm{err}_s(\theta)- \mathbb{E}_{ D^{\phi}_\tau(\boldsymbol{u})}\mathbb{E}_{ D^{\theta}_s(z\mid \boldsymbol{u})}\log D_s^\theta(z \mid \boldsymbol{u}) \ . \label{eq:a1-6}
\end{eqnarray}
We will bound $\mathbf{A}$ and $\mathbf{B}$ next.\vspace{1mm}

\vspace{1mm}
\noindent {\bf \underline{1.~Bounding $\mathbf{A}$.}}~Plugging Eq.~\eqref{eq:a1-2} into Eq.~\eqref{eq:a1-5}, we have
\begin{eqnarray}
\mathbf{A} &=& \mathbb{E}_{D_\tau^\phi(\boldsymbol{u})}\mathbb{E}_{D_s^\theta(z\mid\boldsymbol{u})}\Big[\log D_s^\theta(z\mid \boldsymbol{u})\Big] \ -\  \mathbb{E}_{D_\tau^\phi(\boldsymbol{u})}\mathbb{E}_{D_\tau^\phi(z'\mid\boldsymbol{u})}\Big[\log p_\tau(z'\mid \boldsymbol{u})\Big] \ . \label{eq:a1-7}
\end{eqnarray}
Following Definition~\ref{def:4}, let $\Lambda^*_{\boldsymbol{u}}(z'\mid z) \in C_{\boldsymbol{u}}^\ast$ denote a valid transport map from $D_s^\theta(z\mid \boldsymbol{u})$ to $D_\tau^\phi(z'\mid \boldsymbol{u})$.~That is,
\begin{eqnarray}
\hspace{-26mm}D_\tau^\phi(z'\mid \boldsymbol{u}) &=& \mathbb{E}_{D_s^\theta(z\mid\boldsymbol{u})}\Big[\Lambda_{\boldsymbol{u}}^\ast(z'\mid z)\Big] \ .  \label{eq:a1-8} 
\end{eqnarray}
Plugging Eq.~\eqref{eq:a1-8} into Eq.~\eqref{eq:a1-7}, we can rewrite
\begin{eqnarray}
\mathbf{A} &=& \mathbb{E}_{D_\tau^\phi(\boldsymbol{u})}\mathbb{E}_{D_s^\theta(z\mid\boldsymbol{u})}\Big[\log D_s^\theta(z\mid \boldsymbol{u})\Big] \ -\  \mathbb{E}_{D_\tau^\phi(\boldsymbol{u})}\mathbb{E}_{D_s^\theta(z\mid\boldsymbol{u})}\left[\sum_{z'}\Lambda_{\boldsymbol{u}}^\ast(z'\mid z)\log p_\tau(z'\mid \boldsymbol{u})\right] \ .\label{eq:a1-9}
\end{eqnarray}
Furthermore, we also have
\begin{eqnarray}
\hspace{-6.5mm}-\log p_\tau(z'\mid u) \hspace{-2mm}&=&\hspace{-2mm} -\log\left(\sum_a\Lambda_{\boldsymbol{u}}(z' \mid a) D_s^\theta(a\mid \boldsymbol{u})\right) \ \leq\  -\log \Big(\Lambda_{\boldsymbol{u}}(z' \mid z) D_s^\theta(z\mid \boldsymbol{u})\Big) \nonumber\\
\hspace{-2mm}&=&\hspace{-2mm} -\log \Lambda_{\boldsymbol{u}}(z' \mid z) - \log D_s^\theta(z\mid \boldsymbol{u}) \ .
\label{eq:a1-10}
\end{eqnarray}
for any $z$ and valid transport map $\Lambda_{\boldsymbol{u}}(z' \mid .)\in C_{\boldsymbol{u}}$ from $D_s^\theta(z\mid \boldsymbol{u})$ to $p_\tau(z'\mid \boldsymbol{u})$ as defined in Definition~\ref{def:5}.~Plugging Eq.~\eqref{eq:a1-10} into Eq.~\eqref{eq:a1-9}, we obtain the following upper-bound,
\begin{eqnarray}
\hspace{-20mm}\mathbf{A} &\leq& \mathbb{E}_{D_\tau^\phi(\boldsymbol{u})}\mathbb{E}_{D_s^\theta(z\mid\boldsymbol{u})}\Big[\log D_s^\theta(z\mid \boldsymbol{u})\Big] \nonumber\\
&-& \mathbb{E}_{D_\tau^\phi(\boldsymbol{u})}\mathbb{E}_{D_s^\theta(z\mid\boldsymbol{u})}\left[\sum_{z'}\Lambda_{\boldsymbol{u}}^\ast(z'\mid z)\log \Lambda_{\boldsymbol{u}}(z'\mid z)\right] \ -\ \mathbb{E}_{D_\tau^\phi(\boldsymbol{u})}\mathbb{E}_{D_s^\theta(z\mid\boldsymbol{u})}\left[\sum_{z'}\Lambda_{\boldsymbol{u}}^\ast(z'\mid z)\log D_s^\theta(z\mid \boldsymbol{u})\right]\\
&=& \mathbb{E}_{D_\tau^\phi(\boldsymbol{u})}\mathbb{E}_{D_s^\theta(z\mid\boldsymbol{u})}\Big[\log D_s^\theta(z\mid \boldsymbol{u})\Big]\nonumber\\
&-& \mathbb{E}_{D_\tau^\phi(\boldsymbol{u})}\mathbb{E}_{D_s^\theta(z\mid\boldsymbol{u})}\left[\sum_{z'}\Lambda_{\boldsymbol{u}}^\ast(z'\mid z)\log \Lambda_{\boldsymbol{u}}(z'\mid z)\right] \ -\ 
\mathbb{E}_{D_\tau^\phi(\boldsymbol{u})}\mathbb{E}_{D_s^\theta(z\mid\boldsymbol{u})}\Big[\log D_s^\theta(z\mid \boldsymbol{u})\Big]\nonumber\\
&=&\mathbb{E}_{D_\tau^\phi(\boldsymbol{u})}\mathbb{E}_{D_s^\theta(z\mid\boldsymbol{u})}\left[-\sum_{z'}\Lambda_{\boldsymbol{u}}^\ast(z'\mid z)\log \Lambda_{\boldsymbol{u}}(z'\mid z)\right]\ =\ \mathbb{E}_{D_\tau^\phi(\boldsymbol{u})}\mathbb{E}_{D_s^\theta(z\mid\boldsymbol{u})}\ \mathbb{H}\Big[\Lambda^\ast_{\boldsymbol{u}}(.\mid z), \Lambda_{\boldsymbol{u}}(.\mid z)\Big]\nonumber\\
&=& \mathbb{E}_{D_\tau^\phi(\boldsymbol{u})}\mathbb{E}_{D_s^\theta(z\mid\boldsymbol{u})}\ \mathbb{H}\Big[\Lambda^\ast_{\boldsymbol{u}}(.\mid z)\Big] \ +\ \mathbb{E}_{D_\tau^\phi(\boldsymbol{u})}\ \mathbb{E}_{D_s^\theta(z\mid\boldsymbol{u})}\ \mathbb{KL}\Big[\Lambda^\ast_{\boldsymbol{u}}(.\mid z)\ \|\ \Lambda_{\boldsymbol{u}}(.\mid z)\Big] \ .
\label{eq:a1-11}
\end{eqnarray}
As Eq.~\eqref{eq:a1-11} holds for all choices of $\Lambda^\ast_{\boldsymbol{u}}(z'\mid z) \in C^\ast_{\boldsymbol{u}}$ and $\Lambda_{\boldsymbol{u}}(z'\mid z) \in C_{\boldsymbol{u}}$, we can tighten its right-hand side (RHS) by choosing for each $(\boldsymbol{u})$, $\Lambda^\ast_{\boldsymbol{u}}( \cdot \mid \cdot)\in C_{\boldsymbol{u}}^\ast$ that minimizes $\mathbb{E}_{D^{\theta}_{s}(z| \boldsymbol{u})}\Big[\mathbb{H}[\Lambda_{\boldsymbol{u}}^\ast(.\mid z)] \Big]$ and taking minimum over the remaining choice of $\Lambda_{\boldsymbol{u}}( \cdot \mid \cdot)\in C_{\boldsymbol{u}}$.~This results in a tighten bound below:
\begin{eqnarray}
\hspace{-10mm}\mathbf{A} &\leq& \mathbb{E}_{D^{\phi}_{\tau}(\boldsymbol{u})}\left[\min_{\Lambda_{\boldsymbol{u}}^\ast\in C_{\boldsymbol{u}}^\ast}\mathbb{E}_{D^{\phi}_s(z| \boldsymbol{u})}\Big[\mathbb{H}\Big[\Lambda_{\boldsymbol{u}}^\ast(.\mid z)\Big]\Big]\right] \ +\  \mathbb{E}_{D^{\phi}_{\tau}(\boldsymbol{u})}\left[\min_{\Lambda_{\boldsymbol{u}} \in C_{\boldsymbol{u}}} \mathbb{E}_{D^{\theta}_{s}(z| \boldsymbol{u})}\Big[\mathbb{KL}\Big(\Lambda_{\boldsymbol{u}}^+(.\mid z)\| \Lambda_{\boldsymbol{u}}(.\mid z)\Big)\Big]\right] \ .\label{eq:a1-12}
\end{eqnarray}
where $\Lambda^+_{\boldsymbol{u}}(.\mid .) = \text{argmin}_{\Lambda^\ast_{\boldsymbol{u}}\in C_{\boldsymbol{u}}^\ast} \ \mathbb{E}_{D^{\theta}_{s}(z| \boldsymbol{u})}\Big[\mathbb{H}[\Lambda^\ast_{\boldsymbol{u}}(.\mid z)] \Big]$.~Following the definition of $\mathbf{FLD}$ and $\mathbf{TF}$ in Eqs.~\eqref{eq:FLD-def}-\eqref{eq:TF}, 
\begin{eqnarray}
\mathbf{A} &\leq& \mathbb{E}_{D^{\phi}_{\tau}(\boldsymbol{u})}\Big[\mathbf{FLD}(\boldsymbol{u}) \ +\ \mathbf{TF}(\boldsymbol{u})\Big]  \ , \label{eq:a1-13}  
\end{eqnarray}

\vspace{1mm}
\noindent {\bf \underline{2.~Bounding $\mathbf{B}$.}}~We will now show that $\mathbf{B}$ in Eq.~\eqref{eq:a1-6} is upper-bounded by $\mathbf{FA}(\phi, \theta)$ in Eq.~\eqref{eq:FA-def} to complete the proof.~To see this, note that in practice, $p_s(z\mid\boldsymbol{u})$ often approximates $D_s^\theta(z\mid\boldsymbol{u})$ faithfully.~Exploiting this practical property, we can rewrite
 $\mathbf{B}$ in Eq.~\eqref{eq:a1-6} as
\begin{eqnarray}
\mathbf{B} &=& \mathbb{E}_{D_\tau^\phi(\boldsymbol{u})}\Big[-\mathbb{E}_{D_s^\theta(z\mid \boldsymbol{u})}\big[\log p_s(z\mid \boldsymbol{u})\big]\Big] \ -\  \mathbb{E}_{D_s^\theta(\boldsymbol{u})}\Big[-\mathbb{E}_{D_s^\theta(z\mid\boldsymbol{u})}\big[\log p_s(z\mid \boldsymbol{u})\big]\Big] \\
\hspace{-19mm}&=& \mathbb{E}_{D_\tau^\phi(\boldsymbol{u})}\big[\ell_s(\boldsymbol{u})\big] - \mathbb{E}_{D_s^\theta(\boldsymbol{u})}\big[\ell_s(\boldsymbol{u})\big] \ .\label{eq:a1-14}
\end{eqnarray}
where $\ell_s(\boldsymbol{u}) \triangleq \mathbb{E}_{D_s^\theta(z\mid\boldsymbol{u})}[-\log p_s(z\mid \boldsymbol{u})]$ as previously defined in Definition~\ref{def:6}.~For any cost metric $\delta \in \Delta$ such that $|\ell_s(\boldsymbol{u}_1) - \ell_s(\boldsymbol{u}_2)| \leq \tau_\delta \cdot\delta(\boldsymbol{u}_1, \boldsymbol{u}_2)$, the Kantorovich-Rubinstein duality ascertains that
\begin{eqnarray}
\hspace{-34mm}\mathbf{B} &\leq& \tau_\delta \cdot W_\delta\Big(D_\tau^\phi(\boldsymbol{u}), D_s^\theta(\boldsymbol{u})\Big) \ .\label{eq:a1-15}    
\end{eqnarray}
As this is true for any $\delta \in \Delta$, we can again tighten the right-hand side (RHS) of Eq.~\eqref{eq:a1-15} via taking minimum over the choice of $\delta$.~That is,
\begin{eqnarray}
\mathbf{B}&\leq& \min_{\delta \in \Delta}\Big\{\tau_\delta \cdot W_\delta\Big(D_\tau^\phi(\boldsymbol{u}), D_s^\theta(\boldsymbol{u})\Big)\Big\} \ \ =\ \  \mathbf{FA}(\phi,\theta) \ .\label{eq:a1-16}    
\end{eqnarray}
Plugging Eqs.~\eqref{eq:a1-13} and~\eqref{eq:a1-16} in Eq.~\eqref{eq:a1-4} completes our proof. 

\begin{table*}[htbp]
\centering
\caption{ Comparison of RECRAFT and Baselines on Multiple PDE Tasks (↓ indicates lower is better).~U-Net results for Navier-Stokes and Darcy-Flow are unavailable (due to memory constraints) in the benchmark paper. RECRAFT achieves an average rank of \textbf{1.5} and attains the best performance on 4 out of 8 tasks. Columns \textbf{\#1} and \textbf{\#2} indicate the number of times each method achieves the best and second-best performance, respectively.}\vspace{2mm}
\label{tab:pde-result-expert}
\resizebox{\textwidth}{!}{
\begin{tabular}{|l||llllllll||lll|}
\toprule
\multirow{3}{*}{\textbf{Model}} 
& \begin{tabular}[l]{@{}l@{}}\textbf{Darcy} \\(2D)\end{tabular} 
& \begin{tabular}[l]{@{}l@{}}\textbf{Advection} \\(1D)\end{tabular} 
& \begin{tabular}[l]{@{}l@{}}\textbf{Burgers} \\(1D)\end{tabular} 
& \begin{tabular}[l]{@{}l@{}}\textbf{Diffusion-}\\ \textbf{Sorption} (1D) \end{tabular} 
& \begin{tabular}[l]{@{}l@{}}\textbf{Shallow }\\ \textbf{Water}  (2D)\end{tabular} 
& \begin{tabular}[l]{@{}l@{}}\textbf{Diffusion-}\\ \textbf{Reaction} (2D) \end{tabular} 
& \begin{tabular}[l]{@{}l@{}}\textbf{Diffusion-}\\ \textbf{Reaction} (1D)  \end{tabular} 
& \begin{tabular}[l]{@{}l@{}}\textbf{Navier-}\\ \textbf{Stokes}  (1D) \end{tabular} 
& \multirow{2}{*}{\begin{tabular}[c]{@{}c@{}}\textbf{Avr.}\\\textbf{Rank}\end{tabular}} 
& \multirow{2}{*}{\textbf{\#1}} 
& \multirow{2}{*}{\textbf{\#2}}\\
& nRMSE & nRMSE & nRMSE & nRMSE & nRMSE & nRMSE & nRMSE & nRMSE \\
\midrule
PINN            & $0.18$  & $0.67$   & $0.36$   & $0.15$   & $0.085$  & $0.84$ & $0.84$   & $0.720$  & 3.125 & 0 & 1 \\
FNO             & $0.22$  & $0.011$  & $\textbf{\textcolor{blue}{0.0031}}$ & $1.8\text{\small E-}{3}$ & $\textbf{\textcolor{blue}{4.4\text{\small E-}{3}}}$ & $\textbf{\textcolor{blue}{0.12}}$ & $\textbf{\textcolor{blue}{1.4\text{\small E-}{3}}}$ & $0.068$ & 1.625 & 4 & 3 \\
U-Net           & -     & $1.1$    & $0.99$   & $0.22$   & $0.017$  & $1.6$ & $0.08$   & -     & - & 0 & 0 \\
\midrule
RECRAFT            & $\textbf{\textcolor{blue}{0.079}}$ & $\textbf{\textcolor{blue}{0.0078}}$ & $0.0108$ & $\textbf{\textcolor{blue}{1.6\text{\small E-}{3}}}$ & $5.4\text{\small E-}{3}$ & $0.817$ & $2.8\text{\small E-}{3}$ & $\textbf{\textcolor{blue}{0.050}}$ & $\textbf{\textcolor{blue}{1.500}}$ & 4 & 4 \\
\bottomrule
\end{tabular}
}
\end{table*}

\begin{table*}[t]
\centering
\small
\tiny 
\setlength{\tabcolsep}{2pt}
\caption{ Prediction errors (↓) incurred by the tested methods (with standard deviation) across 10 diverse tasks on NAS-Bench-360.~The reported results of MoNA~\citep{ma2024learningmodalityknowledgealignment} are quoted from the corresponding paper since its source code is not released.~Our method RECRAFT achieves best performance in 8 out 10 tasks.}\vspace{2mm}
\label{tab:complete_nas_results}
\resizebox{1.0\linewidth}{!}{
\begin{tabular}{|l||l|l|l|l|l|l|l|l|l|l|} 
\toprule
\multirow{2}{*}{\textbf{Model}} & \textbf{Darcy} & \textbf{DeepSEA} & \textbf{ECG} & \textbf{CIFAR100} & \textbf{Satellite} & \textbf{Spherical} & \textbf{Ninapro} & \textbf{Cosmic} & \textbf{Psicov} & \textbf{FSD50K} \\
 & Relative $\ell_2$&\text{1-} AUROC   & \text{1-}$\text{F}_1$ score & $\text{0-1 } \text{error } (\%) $ & $\text{0-1 } \text{error } (\%) $ &  $\text{0-1 } \text{error} (\%) $ & $\text{0-1 } \text{error } (\%) $ & \text{1-} AUROC & $\text{MAE}_8$ & \text{1-}mAP\\
\midrule
    Hand-designed   & $8.0\text{E-}{3}\pm 1\text{E-}{3}$ & $0.300\pm 2\text{E-}{2}$ & $0.28\pm 0.01$ & $19.39\pm 0.2$ & $19.80\pm 0.01$  & $67.41\pm0.8$ & $8.74\pm0.9$ & $0.13\pm 1\text{E-}{2}$ & $3.37\pm0.15$ & $0.62\pm4\text{E-}{3}$  \\
    NAS-Bench-360   & $2.6 \text{E-}{2}\pm 1\text{E-}{3}$ & $0.320 \pm 1\text{E-}{2}$ & $0.34\pm0.01$ & $23.39\pm0.03$ & $12.51\pm 0.25$ & $48.23\pm2.5$ & $7.35\pm0.8$ & $0.23\pm 4 \text{E-}{3}$ & $2.95\pm0.14$ & $0.60\pm0.03$ \\
    DASH            & $8.0\text{E-}{3}\pm 2\text{E-}{3}$ & $0.280\pm 1\text{E-}{2}$ & $0.32\pm6\text{E-}{3}$ & $24.37\pm0.83$ & $12.28\pm0.50$ & $71.38\pm0.7$ & $6.63\pm0.4$ & $0.20\pm 5\text{E-}{3}$ & $3.30\pm0.17$ & $0.60\pm0.02$  \\
    Perceiver IO    & $2.4 \text{E-}{2}\pm 1\text{E-}{2}$ & $0.380\pm 4\text{E-}{3}$ & $0.66\pm 0.01$ & $70.04\pm0.4$ & $15.96\pm0.01$ & $82.57\pm0.2$ & $22.4\pm1.5$ & $0.49\pm1\text{E-}{2}$ & $8.10\pm0.05$ & $0.73\pm0.02$ \\
\midrule
NFT         & $7.4\text{E-}{3}\pm 1\text{E-}{4}$ & $0.490\pm 2\text{E-}{2}$   & $0.44\pm0.03$ & $9.74\pm1.21$  & $13.82\pm0.24$ & $55.76\pm2.3$ & $8.35\pm0.4$ & $0.17\pm 2\text{E-}{2}$  & $1.92\pm0.06$ & $0.63\pm0.01$  \\
ORCA       & $7.5\text{E-}{3} \pm 8\text{E-}{5}$  & $0.291\pm 2\text{E-}{3}$  & $0.30\pm 6\text{E-}{3}$ & $7.80\pm0.41$ & $11.63\pm0.20$ & $29.87\pm0.8$ & $7.74\pm0.4$ & $0.15\pm5\text{E-}{3}$  & $1.91\pm0.04$ & $0.56\pm0.01$ \\
PaRE       & $7.4\text{E-}{3}\pm 1\text{E-}{4}$ & $0.286\pm 4\text{E-}{3}$  & $0.28\pm 7\text{E-}{3}$ & $6.70 \pm0.3$ & $11.21\pm0.07$ & $27.04\pm0.7$ & $7.12\pm0.3$ & $0.12\pm 4\text{E-}{3}$  & $\textcolor{blue}{\textbf{0.99}\pm \textbf{0.03}}$ & $\textcolor{blue}{\textbf{0.55} \pm \textbf{0.01}}$ \\
MoNA & $\textbf{\textcolor{blue}{6.8\text{E-}{3}}}$ & 0.280 & $\textbf{\textcolor{blue}{0.27}}$ & $\textbf{\textcolor{blue}{6.48}}$ & $11.13$ & $27.13$ & $7.28$ & $0.121$ & $\textbf{\textcolor{blue}{0.99}}$ & $\textbf{\textcolor{blue}{0.55}}$\\
RECRAFT       & $7.2\text{E-}{3} \pm 7\text{E-}5$& $\textcolor{blue}{\textbf{0.278} \pm \textbf{2}\textbf{E-}{3}}$  & $\textcolor{blue}{\textbf{0.27}\pm \textbf{6}\textbf{E-}{3}}$ & $7.30\pm 0.4$ & $\textcolor{blue}{\textbf{11.11} \pm \textbf{0.04}}$ & $\textcolor{blue}{\textbf{26.41} \pm \textbf{0.7}}$ & $\textcolor{blue}{\textbf{6.60} \pm \textbf{0.3}}$ & $\textcolor{blue}{\textbf{0.11} \pm \textbf{3}\textbf{E-}{3} }$ & $\textcolor{blue}{\textbf{0.99}\pm \textbf{0.02}}$ & $\textcolor{blue}{\textbf{0.55} \pm \textbf{0.01}}$ \\
\bottomrule
\end{tabular}
}
\end{table*}

\section{Tightness of the Generalization Bound in Theorem~\ref{thm:2}}
\label{sec:inequality_effective}
This section evaluates the empirical tightness of the bound in Theorem~\ref{thm:2} which is quoted below for convenience:
\begin{eqnarray}
\mathrm{err}_\tau(\phi) &\leq&  \mathrm{err}_s(\theta) \ \ \ +\ \ \ \textbf{FA}(\phi, \theta)\ \ +\ \  \mathbb{E}_{D^{\phi}_\tau(\boldsymbol{u})}\Big[\textbf{FLD}(\boldsymbol{u}) \ \ \ +\ \ \    \textbf{TF}(\boldsymbol{u}) \Big] \ . 
\end{eqnarray}

This evaluation is conducted with respect to the pre-trained source encoder $\theta$, the target encoder $\phi$ as well as the corresponding target prediction heads $p_\tau(z'\mid \boldsymbol{u})$ which are learned via optimizing the above bound using our proposed algorithm in the main text.~We will show that the bound is sufficiently tight for the learned target encoder $\phi$ which consequently demonstrates that the bound in Theorem~\ref{thm:2} is a sufficiently good surrogate to optimize for the (inaccessible) target generalization loss.

To achieve this, we use the source's and target's \emph{test sets}, which were not used during the pre-training and fine-tuning of the source and target models, to compute the  corresponding target loss $\text{err}_{\tau}(\phi)$ and source loss $\text{err}_{s}(\theta)$.~To compute the bound on the target generalization loss, we calculate the feature alignment (\textbf{FA}) using Eq.~\eqref{eq:FA-loss} and approximate the feature-label distortion (\textbf{FLD}) using the upper bound in Eq.~\eqref{eq:FLD-loss}. We then compute the optimal target fitting term \textbf{TF} at the learned target feature map $\phi$ and prediction head $p_\tau(z'\mid \boldsymbol{u})$ on the target test set via (1) solving for $\Lambda^+_{\boldsymbol{u}}(.\mid.)$ that minimizes the feature-label distortion ({\bf FLD}),
\begin{eqnarray}
\Lambda^+_{\boldsymbol{u}}( \cdot \mid  \cdot) &=& \text{argmin}_{\Lambda^\ast_{\boldsymbol{u}}\in C_{\boldsymbol{u}}^\ast} \mathbb{E}_{D^{\theta}_{s}(z| \boldsymbol{u})} \Big[\ \mathbb{H}\big[\Lambda^\ast_{\boldsymbol{u}}(.\mid z)\big] \Big]   \ ,  
\end{eqnarray}
with respect to the linear constraint in Eq.~\eqref{eq:FLD-constraint}; and (2) leveraging it to equivalently rewrite the {\bf TF} formulation in Definition~\ref{def:5} as the optimal solution for a convex optimization task with linear constraints:
\begin{eqnarray}
\hspace{-3mm}\mathbf{TF}(\boldsymbol{u}) &=& \min_{\Lambda_{\boldsymbol{u}}(.\mid.)} \ \mathbb{E}_{D^{\theta}_{s}(z| \boldsymbol{u})}\Big[\mathbb{KL}\big(\Lambda^{+}_{\boldsymbol{u}}(\cdot| z) || \Lambda_{\boldsymbol{u}}(\cdot | z) \big) \Big] \nonumber\\ \text{ subject to}
\  \ p_{\tau}(z' | \boldsymbol{u}) &=& \mathbb{E}_{D^{\theta}_{s}(z| \boldsymbol{u})}\Big[ \Lambda_{\boldsymbol{u}}(z' | z)\Big] \ \ \text{for each} \ \ z' \in \mathcal{Z}'  \ .
\end{eqnarray}
The above is a direct consequence of the formulation in Definition~\ref{def:5} when we fix a particular choice of the target's feature map $\phi$ and its corresponding prediction head $p_\tau(z'\mid \boldsymbol{u})$.

Importantly, we note that such formulation assumes knowledge of $\phi$ and $p_\tau(z'\mid \boldsymbol{u})$ as well as the test set and therefore cannot be used as an alternative to our proposed algorithm for learning $\phi$ and $p_\tau(z'\mid \boldsymbol{u})$.~Instead, this formulation is an effective probing tool for post-training inspection/evaluation of the tightness of the theoretical bound in Theorem~\ref{thm:2} as it admits a closed-form solution for $\Lambda_{\boldsymbol{u}}$ in terms of the (learned) target feature map $\phi$, prediction head $p_\tau(z'\mid \boldsymbol{u})$, and the corresponding target test data's induced conditional distribution $D_\tau^\phi(z'\mid \boldsymbol{u})$:
\begin{eqnarray}
\Lambda_{\boldsymbol{u}}(z' | z) &=& \Lambda_{\boldsymbol{u}}^{+}(z' | z) \cdot p_{\tau} (z' | \boldsymbol{u})\ /\ D^{\phi}_{\tau}(z'| \boldsymbol{u})\quad\text{and}\quad 
    \textbf{TF}(\boldsymbol{u}) \ \ =\ \ \mathbb{KL}\Big(D^{\phi}_{\tau}(z' | \boldsymbol{u}) \ \|\  p_{\tau}(z' | \boldsymbol{u})\Big)\ . \label{eq:TF_solve}
\end{eqnarray}
{\bf Remark.}~Eq.~\eqref{eq:TF_solve} also provides direct support for the claim in Section~\ref{sec:stage-two} that minimizing the target loss also minimizes the target fitting error (\textbf{TF}). To elaborate, given a feature map $\phi$ and a specified target task, the oracle conditional distribution $D^{\phi}_{\tau}(z' \mid \boldsymbol{u})$ is fixed.~During training phase in Section~\ref{sec:stage-two}, the target predictor $p_{\tau}(z' \mid \boldsymbol{u})$ is trained to minimize the target loss, which, in turn, reduces the Kullback-Leibler (KL) divergence between $p_{\tau}(z' \mid \boldsymbol{u})$ and $D^{\phi}_{\tau}(z' \mid \boldsymbol{u})$.~This corresponds exactly to Eq.~\eqref{eq:TF_solve}.

\begin{figure}[ht]
    \centering
    \includegraphics[width=0.7\textwidth]{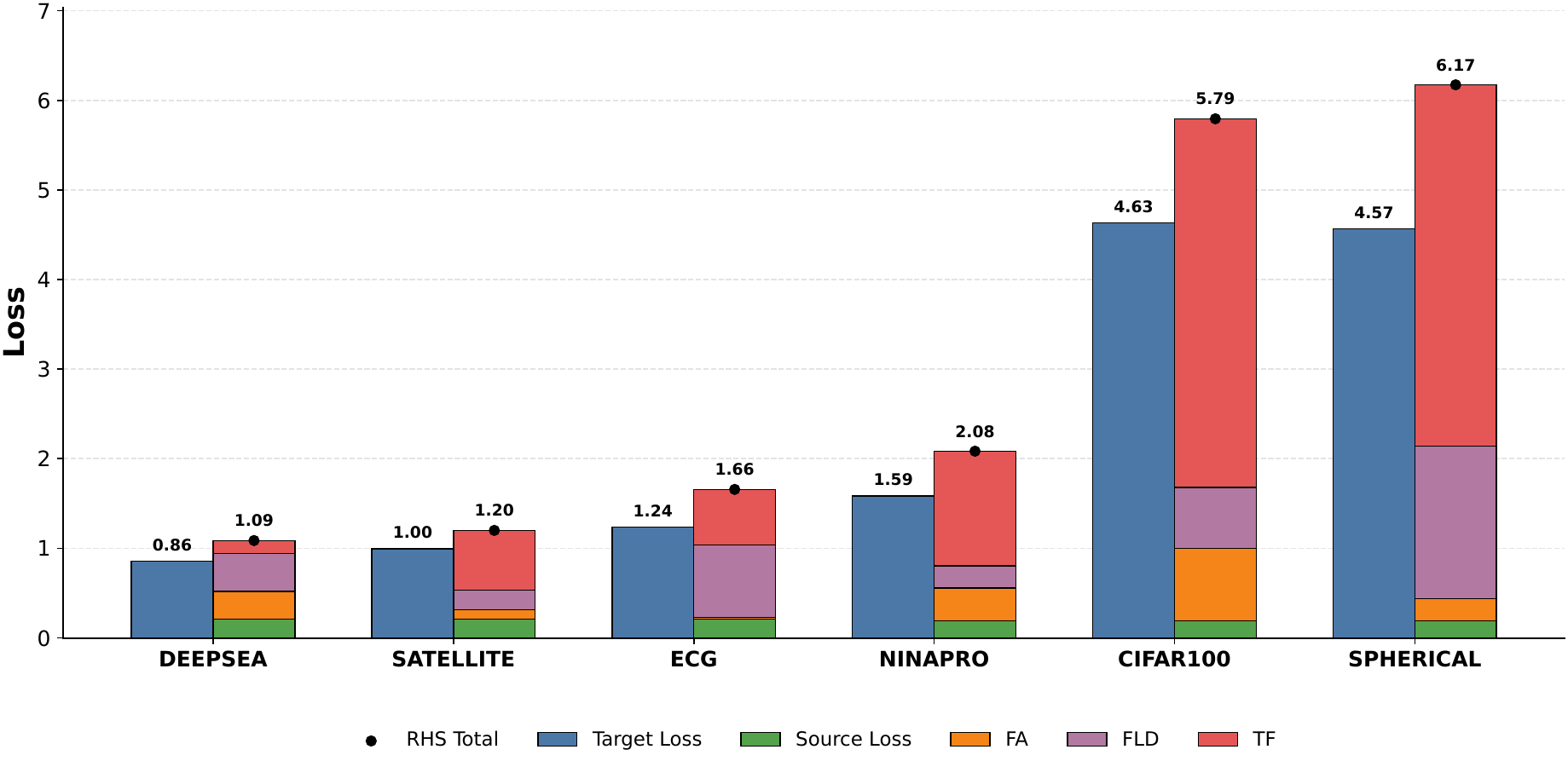}
    \caption{Bar charts illustrating the gap between the target's generalization loss and its upper-bound established in Theorem~\ref{thm:2}.~For each target task, there are two bars representing the target's generalization loss and its upper-bound.~The bar representing the upper-bound is further demarcated into the source's generalization loss (fine-tuning overhead), feature alignment(\textbf{FA}), feature Label distortion(\textbf{FLD}), and target fitting(\textbf{TF}). The bound evaluation is conducted at the learned target's feature map $\phi$ and its corresponding prediction head $p_\tau(z'\mid\boldsymbol{u})$.}
    \label{fig:loss_components_visualize}
\end{figure}

Using the above calculations, we now have access to both the target's generalization loss and its upper-bound based on the source's generalization loss and other key quantities of cross-modal fine-tuning such as feature alignment(\textbf{FA}), feature-label distortion(\textbf{FLD}) and target fitting(\textbf{TF}).~The gap between the target's generalization loss and its upper-bound is visualized in Fig.~\ref{fig:loss_components_visualize} which shows that our error bound is sufficiently tight with small different gap observed consistently across a variety of target datasets.~The averaged gap over all these target tasks is less than 29\%.


\section{Related Works}
\label{sec:full_related_works}
This section summarizes the existing literature on domain adaptation (Appendix~\ref{sec:c1}), in-modal fine-tuning (Appendix~\ref{sec:c2}), and cross-modal fine-tuning (Appendix~\ref{sec:c3}).

\subsection{Domain Adaptation} 
\label{sec:c1}

Domain adaptation is a form of transductive transfer learning in which the inputs and outputs of the source and target tasks are identical, but their corresponding data distributions are different~\citep{pan2009survey, wang2018deep}. Its main focus is on transferring knowledge within the same task across different data distributions (i.e., domains) assuming shared input/output spaces. It also assumes that target inputs are accessible during training (often in the absence of target labels) to enable distribution alignment in a transductive setting. Furthermore, prior to the advent of foundation models (FMs), algorithmic designs for domain adaptation were tailored to specific task (i.e., input/output spaces) and source model's architecture, limiting their applicability to other scenarios. In contrast, fine-tuning in the FM paradigm enables model-agnostic adaptation procedures that operate without assuming specific forms of input/output or requiring access to target data during training. While some recent efforts in domain adaptation have relaxed assumptions on shared input/output spaces~\citep{NghiaICML20,NghiaICML21,NghiaAAAI24,yin2024domainadaptationcauchyschwarzdivergence}, they still require source and target inputs to originate from the same modality (e.g., images, text), which restricts their use in more general cross-modal transfer settings.~Other recent approaches have considered text-to-image transfer but require access to the (unlabeled) target data during training, are restricted to the same task, use a specific model architecture catering to text and image modalities~\citep{Yao2019HeterogeneousDA,Li2020SimultaneousSA}.~This is different from the broader scheme of cross-modal fine-tuning which aims to enable transfer across different tasks and unseen modalities via model-agnostic procedures.

\subsection{In-Modal Fine-Tuning of FMs}
\label{sec:c2}

The recent advent of large pre-trained or foundation models (FMs) has enabled flexible knowledge transfer to a wide range of downstream tasks under a unified, task-agnostic paradigm. This is commonly known as fine-tuning, which is model-agnostic and does not require shared input or output spaces. Such transfer is feasible because these FMs are trained on broad, diverse corpora (e.g., text, image, or audio) that often overlap semantically or structurally with the content of downstream datasets for in-modal tasks. As a result, fine-tuning has been successfully applied across domains such as vision~\citep{kirillov2023segment, liu2021swin}, video understanding~\citep{bertasius2021space}), language~\citep{bo2021xtune, yang2022enhancing, ma2023language}, and speech~\citep{radford2022robustspeechrecognitionlargescale, li-etal-2021-multilingual}.~There are also multi-modal FMs~\citep{Radford2021LearningTV,Alayrac2022FlamingoAV,Kim2021ViLTVT} which were pre-trained to learn the embeddings of multiple modalities together but existing approaches to facilitate knowledge transfer from these models are still restricted to within the pre-trained modalities.~Extending fine-tuning to cross-modal scenarios with unseen data and downstream tasks (e.g., fine-tuning vision/language FMs to solve complex physics problems) are however less explored as discussed next.

\subsection{Cross-Modal Fine-Tuning of FMs}
\label{sec:c3}

Early attempts in cross-modal fine-tuning have focused on transferring language models to other modalities such as vision~\citep{Kiela2019SupervisedMB,tan2019lxmert,gu2022vision}, DNA/protein sequences~\citep{nguyen2023hyenadnalongrangegenomicsequence,lin2023evolutionary,jumper2021highly}.~These provide initial evidence of the cross-modal transferability of FMs but their designs are nonetheless hand-tailored to specific target tasks and modalities rather than for general-purpose~\citep{shen2023orca}.~More recently, a few general-purpose cross-modal fine-tuning frameworks have emerged with remarkable successes in finetuning existing vision~\citep{liu2021swin} or language~\citep{Liu2019RoBERTaAR} FMs to solve a broad and diverse set of unseen tasks and data modalities~\citep{shen2023orca,cai2024enhancingcrossmodalfinetuninggradually,ma2024learningmodalityknowledgealignment}.~In particular, ORCA~\citep{shen2023orca} aligns source and target representation distributions using optimal transport before fine-tuning the entire network.~PARE~\citep{cai2024enhancingcrossmodalfinetuninggradually} alternatively introduces a gating mechanism to integrate source and target features during fine-tuning.~MoNA~\citep{ma2024learningmodalityknowledgealignment} addresses modality representation alignment via a bi-level optimization framework: the inner loop learns the optimal target predictor for a given embedder, while the outer loop updates the embedder to align the source feature-label semantics with the combined representation and prediction under this predictor. This help reduces the misalignment between source's and target's feature-label structures while fitting to the target data.~However, as noted in Section~\ref{sec:intro}, these approaches largely adopt heuristic methods combining distributional feature alignment and target fitting to facilitate cross-modal knowledge transfer.~Overall, these approaches lack a principled means to assess the impact of their heuristic strategies on the generalized target performance.

\section{Technical Background on Wasserstein Distance}
As the core technical block of our theoretical analysis is built on the Wasserstein distance, we provide a succinct definition of this distance (Appendix~\ref{sec:wasserstein}) and its computation (Appendix~\ref{sec:compute_OT}) below.

\subsection{Wasserstein Distance}
\label{sec:wasserstein}
The Wasserstein $p$-distance is a fundamental metric in the theory of optimal transport.~It is used to measure the distance between probability measures defined on a metric space.~Intuitively, it is the cost of transporting mass from one distribution to another with respect to a cost metric $\delta(\boldsymbol{u}_1,\boldsymbol{u}_2)$ measuring the distance between two locations.~This concept is widely applied in fields such as probability theory, machine learning, and partial differential equations \citep{villani2008optimal}.
Let $\big(\mathcal{U}, \delta \big)$ be a metric space that is a Polish space. For $p \in [1, + \infty]$, the Wasserstein-$p$ distance  between two probability measures $\mu$ and $\nu$ on 
space $\mathcal{U}$ is: 
\begin{eqnarray}
W_{\delta}^{p}\big(\mu, \nu\big) &=& \inf_{\pi \in \Pi(\mu, \nu)} \Big( \mathbb{E}_{(\boldsymbol{u}_1, \boldsymbol{u}_2) \sim \pi} \big[\delta(\boldsymbol{u}_1, \boldsymbol{u}_2)^{p} \big]\Big)^{\frac{1}{p}}
\end{eqnarray}
For example, we denote the Wasserstein-$1$ distance between two probability measures $\mu$ and $\nu$ on $\big(\mathcal{U}, \delta \big)$ as: 
\begin{eqnarray}
W_{\delta}(\mu, \nu) &=& \inf_{\pi \in \Pi(\mu, \nu)} \Big( \mathbb{E}_{(\boldsymbol{u}_1, \boldsymbol{u}_2) \sim \pi} \big[\delta(\boldsymbol{u}_1, \boldsymbol{u}_2)  \big]\Big)  \\
&=& \inf_{\pi \in \Pi(\mu, \nu)} \int_{\mathcal{U}} \int_{\mathcal{U}} \delta(\boldsymbol{u}_1, \boldsymbol{u}_2) \pi(\boldsymbol{u}_1, \boldsymbol{u}_2) d \boldsymbol{u}_1 d \boldsymbol{u}_2\ .
\end{eqnarray}
A key property of the Wasserstein-1 distance is the Kantorovich duality theorem, which provides a dual formulation of the distance in terms of Lipschitz functions.~This duality is particularly useful in applications ranging from machine learning to functional analysis~\citep{villani2008optimal} and is stated below.~For any \(\tau_{\delta} > 0\), let \(\text{Lip}_{\tau_{\delta}}\) denote the set of functions \(g: \mathcal{U} \to \mathbb{R}\) such that
\[
\big|g(\boldsymbol{u}_1) - g(\boldsymbol{u}_2)\big| \ \ \leq\ \  \tau_{\delta} \delta\big(\boldsymbol{u}_1, \boldsymbol{u}_2\big)
\]
for all \(\boldsymbol{u}_1, \boldsymbol{u}_2 \in \mathcal{U}\). Then, the Wasserstein-1 distance is given by
\begin{eqnarray}
W_{\delta}(\mu, \nu) &=& \frac{1}{\tau_{\delta}} \sup_{f \in \text{Lip}_{\tau_{\delta}}} \Big( \mathbb{E}_{\boldsymbol{u}_1 \sim \mu}\Big[g(\boldsymbol{u}_1)\Big] \ \ -\ \  \mathbb{E}_{\boldsymbol{u}_2 \sim \nu}\Big[g(\boldsymbol{u}_2)\Big] \Big)\ .
\end{eqnarray}
\subsection{Computation of Wasserstein Distance}
\label{sec:compute_OT}

Computing the  Wasserstein distance is achieved via solving a linear program with a computational complexity of \(O(n^3)\)~\citep{peyré2020computationaloptimaltransport}.~This is however impractical for large datasets.~To address this, we adopt entropic regularization to introduce a penalty term \(\epsilon H(\pi) = -\epsilon \mathbb{E}_{\pi}[\log(\pi)]\) with $\epsilon>0$ which leads to the following augmented optimization:
\begin{eqnarray}
\pi_\epsilon &=& \arg\min_{\pi \in \Pi(\mu, \nu)} \Big( \mathbb{E}_{(\boldsymbol{u}_1, \boldsymbol{u}_2) \sim \pi} \big[\delta(\boldsymbol{u}_1, \boldsymbol{u}_2)  \big] \ +\   \epsilon H(\pi)\Big)  \ .   
\end{eqnarray}
This regularized problem is solved efficiently using the Sinkhorn algorithm, which iteratively scales the transport plan \(\pi_\epsilon = \text{diag}(u) K \text{diag}(v)\), where \(u, v\) are scaling vectors updated to satisfy the marginal constraints.

The algorithm has an overall complexity of \(O(n^2)\) which offers significant speedup for large-scale problems~\citep{peyré2020computationaloptimaltransport}.~For a practical implementation, we utilize the Sinkhorn algorithm as provided by the Python Optimal Transport (POT) library and use its default regularization parameter \(\epsilon = 0.1\) to ensure a balance between computational efficiency and accuracy.

\section{Additional Experimental Results and Ablation Studies}
\label{sec:addtional_result}
This section provides additional experimental results to supplement the main text results.~These results include: 
(i) the performance of RECRAFT compared to specialized physic-informed methods such as PINN~\citep{RAISSI2019686} and FNO~\citep{li2021fourier}, as well as a baseline U-Net method~\citep{10.1007/978-3-319-24574-4_28} on PDEBench as shown in Table~\ref{tab:pde-result-expert}; (ii) experiment results on NAS-Bench-360~\citep{tu2022nasbench} with additional details on error bars for each task, as shown in Table~\ref{tab:complete_nas_results}; (iii) experiment results on PDEBench with additional details on error bars for each task, as shown in Table \ref{tab:complete_pde-result} and Table~\ref{tab:complete_pde-result_cross}. 

We also provide the ablations across several tasks on NAS-Bench-360~\citep{tu2022nasbench} isolating contributions: NFT vs. FA-only vs. RECRAFT, as shown in Table~\ref{tab:complete_nas_FA}. RECRAFT achieves the best performance across all tasks and highlights the contribution of feature-label distortion(FLD) loss. 

\begin{figure}[ht]
    \centering
    \includegraphics[width=0.98\textwidth]{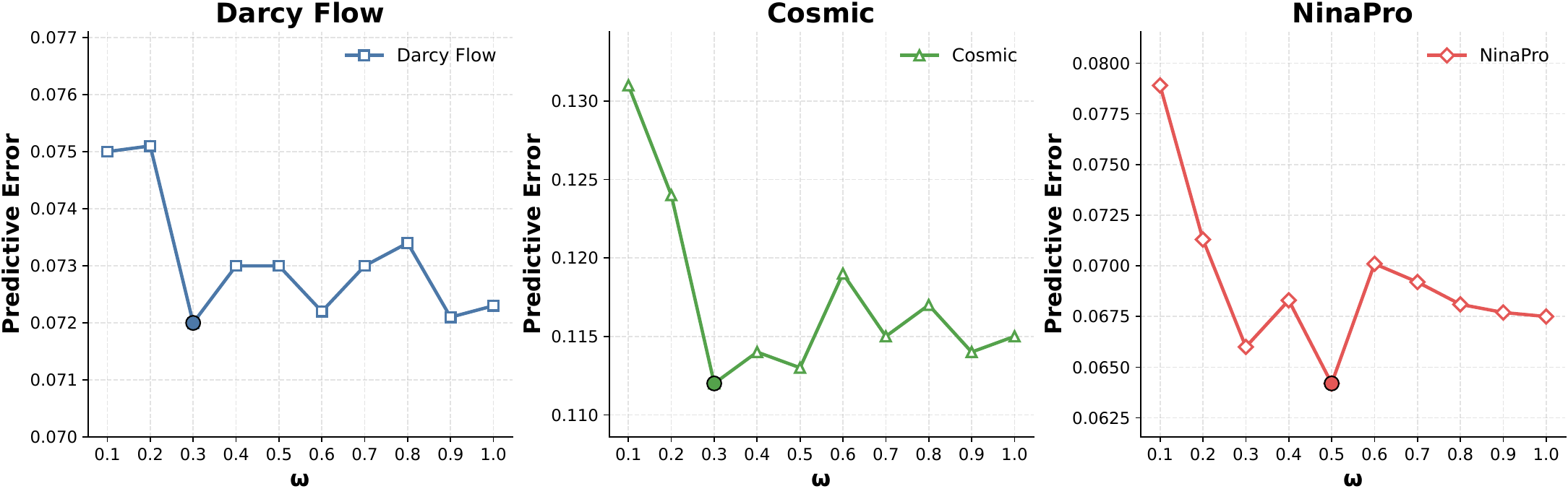}
    \caption{\small Predictive error (↓) of Darcy Flow, Cosmic and Ninapro across various values of $\omega$, which achieves the balance between minimizing Feature Alignment(\textbf{FA}) and Feature Label Distortion(\textbf{FLD}).}
    \label{fig:omega_sensitivity}
\end{figure}

To assess the sensitivity of $\omega$, which ensures the Lipschitz continuity of the loss function $\ell_{s}$ on the source dataset and balances the minimization of feature alignment (\textbf{FA}) and feature-label distortion (\textbf{FLD}), we evaluate the model's performance across different target datasets using varying values of $\omega$.
Fig.~\ref{fig:omega_sensitivity} illustrates the sensitivity of $\omega$ across various target datasets.~The results suggest that setting $\omega$ within the range of approximately 0.3 to 0.5 achieves an optimal balance between \textbf{FA} and \textbf{FLD} during optimization.
\begin{table*}[htbp]
\centering
\caption{Predictive error (↓) (with standard deviation) achieved by our proposed method RECRAFT and other specialized physic-informed baselines across multiple PDE tasks.~U-Net results for Naiver-Stokes and Darcy Flow are missing because the benchmark paper~\citep{Takamoto2022PDEBENCHAE} does not evaluate them (due to memory issues).}\vspace{2mm}
\label{tab:complete_pde-result}

\resizebox{\textwidth}{!}{
\begin{tabular}{|l||l|l|l|l|l|l|l|l|}
\toprule
\textbf{Model} & \textbf{Darcy} & \textbf{Advection} & \textbf{Burgers} & \textbf{Diff.Sorp} & \textbf{S.Water} & \textbf{Diff.Reac}(2D) & \textbf{Diff.Reac}(1D) & \textbf{Navier-Stokes}  \\
 & nRMSE & nRMSE& nRMSE& nRMSE& nRMSE& nRMSE& nRMSE& nRMSE\\
\midrule
PINN            & $0.180\pm2\text{E-}{3}$  & $0.67\pm0.03$   & $0.36\pm0.03$   & $0.15\pm0.03$   & $0.085\pm3\text{E-}{3}$  & $0.840\pm0.01$ & $0.840\pm0.01$   & $0.720\pm 5\text{E-}{3}$   \\
FNO             & $0.220\pm3\text{E-}{3}$  & $0.011\pm1\text{E-}{3}$  & $\textcolor{blue}{\textbf{0.003}1\pm \textbf{5}\text{E-}{5}}$ & $1.8\text{E-}{3}\pm4\text{E-}{5}$ & $\textcolor{blue}{\textbf{4.4} \text{E-}{3}\pm \textbf{3} \text{E-}{5}}$ & $\textcolor{blue}{\textbf{0.120} \pm \textbf{4}\text{E-}{4}}$ & $\textcolor{blue}{\textbf{1.4} \text{E-}{3}\pm \textbf{1}\text{E-}{4}}$ & $0.068\pm2\text{E-}{3}$  \\
U-Net           & -     & $1.1\pm0.21$    & $0.99\pm0.02$   & $0.22\pm0.02$   & $0.017\pm2\text{E-}{3}$  & $1.6\pm 7\text{E-}{3}$ & $0.08\pm 4\text{E-}{3}$   & -      \\
\midrule

RECRAFT            & $\textcolor{blue}{\textbf{0.079}\pm \textbf{1}\textbf{E-}{3}}$ & $\textcolor{blue}{\textbf{0.0078} \pm \textbf{4}\text{E-}{4}}$ & $0.0108\pm3\text{E-}{4}$ & $\textcolor{blue}{\textbf{1.6}\textbf{E-}{3}\pm3\textbf{E-}{4}}$ & $5.4\text{E-}{3}\pm5\text{E-}{5}$ & $0.817\pm3\text{E-}{3}$ & $2.8\text{E-}{3}\pm2\text{E-}{4}$ & $\textcolor{blue}{\textbf{0.050} \pm \textbf{3}\textbf{E-}{3}}$  \\
\bottomrule
\end{tabular}
}
\end{table*}

\begin{table*}[h]
\centering
\caption{ \small Predictive error (↓) (with standard deviation) achieved by our proposed method RECRAFT and other state-of-the-art (SOTA) cross-modal fine-tuning baselines across multiple PDE tasks in PDEBench~\citep{Takamoto2022PDEBENCHAE}.~The reported results of MoNA~\citep{ma2024learningmodalityknowledgealignment} are sourced from the corresponding paper due to its unavailable code.}\vspace{2mm}
\label{tab:complete_pde-result_cross}

\resizebox{\textwidth}{!}{
\begin{tabular}{|l||l|l|l|l|l|l|l|l|}
\toprule
\textbf{Model} & \textbf{Darcy} & \textbf{Advection} & \textbf{Burgers} & \textbf{Diff.Sorp} & \textbf{S.Water} & \textbf{Diff.Reac}(2D) & \textbf{Diff.Reac}(1D) & \textbf{Navier-Stokes}  \\
 & nRMSE & nRMSE& nRMSE& nRMSE& nRMSE& nRMSE& nRMSE& nRMSE\\
\midrule

NFT             & $0.085\pm 4\text{E-}{3}$ & $0.0140\pm2\text{E-}{3}$  & $0.0130\pm4\text{E-}{4}$  & $3.1\text{E-}{3}\pm7\text{E-}{5}$ & $6.1\text{E-}{3}\pm4\text{E-}{5}$ & $0.830\pm4\text{E-}{3}$  & $9.2\text{E-}{3}\pm2\text{E-}{3}$ & $0.863\pm 2\text{E-}{2}$  \\
ORCA            & $0.081\pm 1\text{E-}{3}$ & $0.0098\pm2\text{E-}{4}$ & $0.0120\pm4\text{E-}{4}$ & $1.8\text{E-}{3}\pm2\text{E-}{4}$ & $6.0\text{E-}{3}\pm5\text{E-}{5}$ & $0.820\pm3\text{E-}{3}$ & $3.2\text{E-}{3}\pm2\text{E-}{4}$ & $0.066\pm2\text{E-}{3}$  \\
PARE            & $0.081\pm 1\text{E-}{3}$ & $\textcolor{blue}{\textbf{0.0032} \pm \textbf{4}\text{E-}{4}}$ & $0.0114\pm3\text{E-}{4}$ & $1.9\text{E-}{3}\pm3\text{E-}{4}$ & $5.9\text{E-}{3}\pm5\text{E-}{5}$ & $0.820\pm5\text{E-}{3}$ & $2.9\text{E-}{3}\pm3\text{E-}{4}$ & $0.068\pm3\text{E-}{3}$  \\
MoNA & $\textbf{\textcolor{blue}{0.079}}$ & $0.0088$ & $0.0114$ & $\textcolor{blue}{\textbf{1.6}\textbf{E-}{3}}$ & $5.7\text{E-}{3}$ & $0.818$ & $\textcolor{blue}{\textbf{2.8}\text{E-}{3}}$& $0.054$\\
RECRAFT            & $\textcolor{blue}{\textbf{0.079}\pm \textbf{1}\textbf{E-}{3}}$ & $0.0078\pm4\text{E-}{4}$ & $\textcolor{blue}{\textbf{0.0108} \pm \textbf{3}\text{E-}{4}}$ & $\textcolor{blue}{\textbf{1.6}\textbf{E-}{3}\pm3\textbf{E-}{4}}$ & $\textcolor{blue}{\textbf{5.4}\text{E-}{3}\pm \textbf{5} \text{E-}{5}}$ & $\textcolor{blue}{\textbf{0.817} \pm \textbf{3} \text{E-}{3}}$ & $\textcolor{blue}{\textbf{2.8}\text{E-}{3}\pm \textbf{2}\text{E-}{4}}$ & $\textcolor{blue}{\textbf{0.050} \pm \textbf{3}\textbf{E-}{3}}$  \\
\bottomrule
\end{tabular}
}
\end{table*}
\section{Practical Source Re-calibration to Ensure Lipschitz Constraint with $\ell_2$ Metric}
\label{sec:FA_assumption_enforce}
\begin{wrapfigure}{r}{0.5\textwidth}
    \centering
    \vspace{-6mm}
\includegraphics[width=0.5\textwidth]{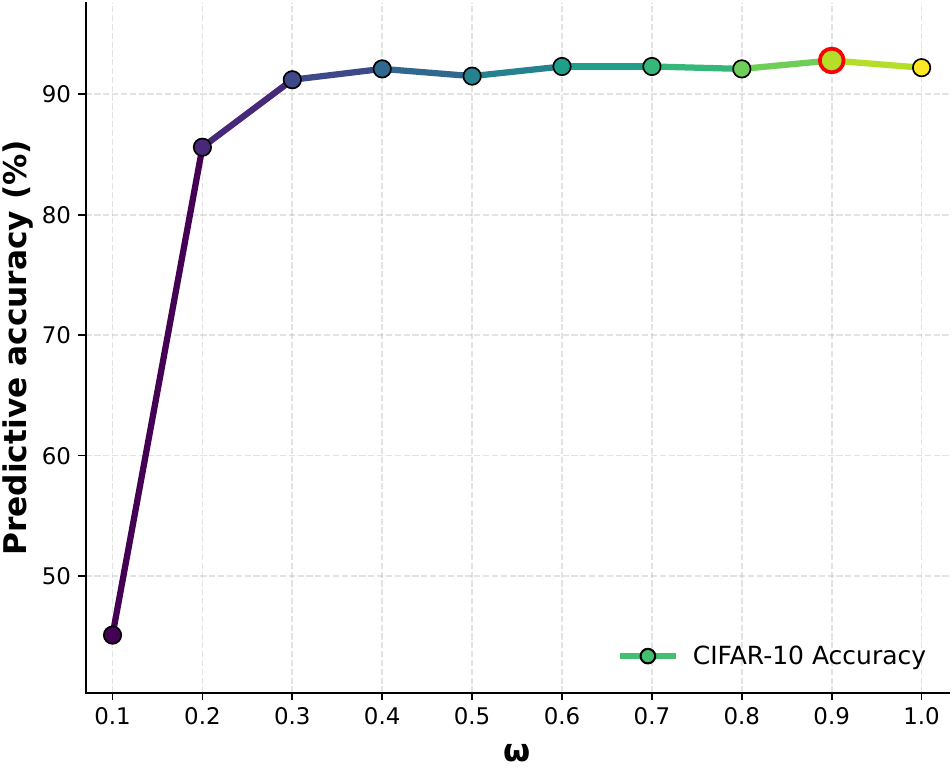} 
    \vspace{-10mm}
    \caption{\small Performance of the source model on the (proxy) CIFAR-10 dataset with re-calibrated prediction head to satisfy the Lipschitz constraint in Definition~\ref{def:6} with constant $\tau_\delta = \omega$ where $\omega$ varies in $[0.1, 1.0]$.}\vspace{2mm}
    \label{tab:tau_selection}
\end{wrapfigure}
This section provides a concrete example of how the source's prediction map can be recalibrated to ensure Lipschitz constraint  with a practical choice of $\delta$.~Note that the theoretical bound in Theorem~\ref{thm:2} holds with any metric, we can then choose $\delta$ to be a Euclidean distance, i.e., $\delta \equiv \ell_2$.~Given the choice, we can now aim to minimize $\tau_\delta = \omega$ via recalibrating the source's prediction map.~This is achieved via finding a smallest value of $\omega$ for which the source's prediction map can be refitted so that $\ell_s(\boldsymbol{u}): = -\mathbb{E}_{D^{\theta}_s(z \mid \boldsymbol{u})} [\log(p_s(z \mid \boldsymbol{u})) ]$ is $O(\omega)$-Lipschitz with respect to the Euclidean metric $\delta \equiv \ell_2$ while preserving performance on the proxy dataset.~To do this, we view $\omega$ as a hyper-parameter and run ablation experiments on the proxy dataset with different choices of $\omega$ over a pre-defined range $[0.1, 1.0]$ to determine its optimal choice.~For each $\omega$, we re-calibrate the source's prediction map via\vspace{-2mm}
\begin{eqnarray}
\hspace{-2mm}\gamma(\omega) \hspace{-2.5mm}&=&\hspace{-2.5mm} \underset{\gamma}{\text{argmin}}\hspace{0.5mm}\underset{\boldsymbol{x} \sim P_s}{\mathbb{E}} \hspace{-1mm}\max \Big(0, \|\nabla_{\boldsymbol{u}} \ell_s(\theta(\boldsymbol{x}); \gamma)\|_2 - \omega \Big)\nonumber
\end{eqnarray}

\vspace{-6mm}
with $\gamma$ denotes the last layer of $p_s(z\mid \boldsymbol{u})$.~Once the re-calibrated $\gamma(\omega)$ has been computed, we re-evaluate the source performance to determine whether $\gamma(\omega)$ preserves performance and $\omega$ can be further reduced.

Our experiments show that smaller values of \(\omega\) indeed lead to larger source loss values as they impose more constraints on the prediction map.~Based on these ablation results, we selected \(\omega = 0.3\) for 2D tasks; and \(\omega = 0.5\) for 1D tasks.~These are the minimum values of $\omega$ for which the re-calibrated prediction map preserves the source's model on the proxy dataset.~To visualize this, we present Fig.~\ref{tab:tau_selection} which plots the re-calibrated (source) model's performance on the proxy CIFAR-10 dataset versus Lipschitz threshold $\tau_\delta \triangleq \omega \in [0.1, 1.0]$ (for 2D tasks).

\section{RECRAFT algorithm}
\label{sec:pseudocode}
For better clarity, this section provides the pseudocode for our algorithm RECRAFT previously described in Section~\ref{sec:algo}.~To summarize, RECRAFT adopts a two-stage approach that decomposes the bound minimization into (1) finding a feature map $\phi$ that minimizes the semantic gap $\textbf{FA}(\phi,\theta) + \mathbb{E}_{D^{\phi}_{\tau}(\boldsymbol{u})}[\textbf{FLD}(\boldsymbol{u})]$ between the source and target tasks, thus maximizing transferability (Section~\ref{sec:stage-one}); and (2) learning a predictor $p_\tau(z'\mid \boldsymbol{u})$ based on the learned $\phi$ (Section~\ref{sec:stage-two}) via minimizing $\mathbb{E}_{D^{\phi}_{\tau}(\boldsymbol{u})}[\textbf{TF}(\boldsymbol{u})]$.~A pseudocode of RECRAFT is detailed in Algorithm~\ref{alg:recraft}. 

\begin{table*}[t]
\centering
\small
\tiny 
\setlength{\tabcolsep}{2pt}
\caption{\small Prediction errors (↓) incurred by NFT(naive fine-tuning),FA(only optimize feature alignment) and RECRAFT (with standard deviation) across 10 diverse tasks on NAS-Bench-360.~Our method RECRAFT achieves best performance all tasks, highlight the effectiveness of minimizing FLD.}\vspace{2mm}
\label{tab:complete_nas_FA}
\resizebox{1.0\linewidth}{!}{
\begin{tabular}{|l||l|l|l|l|l|l|l|l|l|l|} 
\toprule
\multirow{2}{*}{\textbf{Model}} & \textbf{Darcy} & \textbf{DeepSEA} & \textbf{ECG} & \textbf{CIFAR100} & \textbf{Satellite} & \textbf{Spherical} & \textbf{Ninapro} & \textbf{Cosmic} & \textbf{Psicov} & \textbf{FSD50K} \\
 & Relative $\ell_2$&\text{1-} AUROC   & \text{1-}$\text{F}_1$ score & $\text{0-1 } \text{error } (\%) $ & $\text{0-1 } \text{error } (\%) $ &  $\text{0-1 } \text{error} (\%) $ & $\text{0-1 } \text{error } (\%) $ & \text{1-} AUROC & $\text{MAE}_8$ & \text{1-}mAP\\
\midrule

NFT         & $7.4\text{E-}{3}\pm 1\text{E-}{4}$ & $0.490\pm 2\text{E-}{2}$   & $0.44\pm0.03$ & $9.74\pm1.21$  & $13.82\pm0.24$ & $55.76\pm2.3$ & $8.35\pm0.4$ & $0.17\pm 2\text{E-}{2}$  & $1.92\pm0.06$ & $0.63\pm0.01$  \\
FA       & $7.3\text{E-}{3} \pm 8\text{E-}{5}$  & $0.290\pm 2\text{E-}{3}$  & $0.29\pm 6\text{E-}{3}$ & $7.80\pm0.41$ & $11.61\pm0.20$ & $29.85\pm0.8$ & $7.73\pm0.4$ & $0.16\pm5\text{E-}{3}$  & $1.92\pm0.04$ & $0.57\pm0.01$ \\

Ours       & $\textcolor{blue}{\textbf{7.2}\text{E-}{3} \pm \textbf{7}\text{E-}5}$& $\textcolor{blue}{\textbf{0.278} \pm \textbf{2}\textbf{E-}{3}}$  & $\textcolor{blue}{\textbf{0.27}\pm \textbf{6}\textbf{E-}{3}}$ & $\textcolor{blue}{\textbf{7.30}\pm \textbf{0.4}}$ & $\textcolor{blue}{\textbf{11.11} \pm \textbf{0.04}}$ & $\textcolor{blue}{\textbf{26.41} \pm \textbf{0.7}}$ & $\textcolor{blue}{\textbf{6.60} \pm \textbf{0.3}}$ & $\textcolor{blue}{\textbf{0.11} \pm \textbf{3}\textbf{E-}{3} }$ & $\textcolor{blue}{\textbf{0.99}\pm \textbf{0.02}}$ & $\textcolor{blue}{\textbf{0.55} \pm \textbf{0.01}}$ \\
\bottomrule
\end{tabular}
}
\end{table*}

\begin{algorithm}
\caption{{\bf RECRAFT} Algorithm}
\label{alg:recraft}
\begin{algorithmic}[1]
\STATE \textbf{Input:} Pre-trained model $M_s \triangleq(\theta, p_s(z \mid \theta(\boldsymbol{x})))$, proxy dataset $(\boldsymbol{X}^s, \boldsymbol{z}^s)\sim D_s(\boldsymbol{x},z)$, target dataset $(\boldsymbol{X}^\tau, \boldsymbol{z}^\tau)\sim D_\tau(\boldsymbol{x}', z')$, number of epochs $n_0$, $n_1$, $n_2$.
\STATE \textbf{Output:} The target model $M_\tau \triangleq (\phi, p_\tau(z' \mid \phi(\boldsymbol{x}')))$.

\STATE \textit{\textbf{Stage 1: Learning Feature Map} (Section~\ref{sec:stage-one})}
\STATE Initialize the target embedder $\phi$.
\FOR{$epoch = 1$ to $n_1$}
    \STATE Compute  $L_{\textbf{FA}}(\phi)$ using Eq.~\ref{eq:FA-loss}.
    \STATE Update $\phi$ by minimizing  $L_{\textbf{FA}}(\phi)$. 
\ENDFOR
\FOR{$epoch = 1$ to $n_2$}
    \STATE Compute $L_{\textbf{FLD}}(\phi)$ using Eq.~\ref{eq:FLD-loss}.
    \STATE Update $\phi$ by minimizing $L_{\textbf{FLD}}(\phi)$.
\ENDFOR

\STATE \textit{\textbf{Stage 2: Learning Target Predictor} (Section~\ref{sec:stage-two})}
\STATE Frozen $\phi$, initialize $p_\tau(z' \mid \boldsymbol{u})$ from $p_s(z \mid \boldsymbol{u})$.
\FOR{$epoch = 1$ to $n_0$}
    \STATE Compute $L_{\textbf{TF}}$ using Eq.~\ref{target_loss}.
    \STATE Update $p_\tau$ by minimizing $L_{\textbf{TF}}$.
\ENDFOR

\STATE \textbf{Return} $(\phi, p_\tau(z' \mid \boldsymbol{u}))$
\end{algorithmic}
\end{algorithm}
\section{Hyperparameters \& Implementation Details}
\label{appdendix:hyperparams}
To ensure fair comparisons, we adopt the same hyperparameters as those used in \textbf{ORCA}~\citep{shen2023orca} for model fine-tuning.~These specific parameter settings are shown in Table~\ref{Nas_hyper} for Nas-Bench-360~\citep{tu2022nasbench} and Table~\ref{pde_hyper} for PDEBench~\citep{Takamoto2022PDEBENCHAE}.~For detailed implementation, we construct the target model \( M_\tau \triangleq (\phi, p_\tau(z' \mid \phi(\boldsymbol{x}'))) \) by adapting a pre-trained foundation model \( M_s \triangleq (\theta, p_s(z \mid \theta(\boldsymbol{x}))) \) which is selected to be a Swin Transformer~\citep{liu2021swin} for 2D tasks; and a RoBERTa~\citep{Liu2019RoBERTaAR} for 1D tasks. 
To compute the FLD for a regression task, we discretize the continuous label space into 10 distinct classes, consistent with the hyperparameter settings used in ORCA~\citep{shen2023orca}. 

All experiments are run on an NVIDIA RTX 2080 GPU and results are averaged over 5 independent runs. Our code repository can be found at 
\url{https://github.com/khiembk/RECRAFT}.

\begin{table*}[htbp]
\centering
\caption{Hyperparameter configurations used for the 10 tasks on NAS-Bench-360~\citep{tu2022nasbench}.~{\em FA epoch} and {\em FLD epoch} denote the number of training epochs used to minimize the feature alignment (\textbf{FA}) and feature-label distortion (\textbf{FLD}).~The optimizers used in our experiments include SGD, Adam~\citep{kingma2017adammethodstochasticoptimization} and AdamW~\citep{loshchilov2017fixing}.}\vspace{2mm}
\label{Nas_hyper}
\resizebox{\textwidth}{!}{
\begin{tabular}{|l|l|l|l|l|l|l|l|l|l|l|}
\toprule
\textbf{Hyperparameter} & \textbf{CIFAR100} & \textbf{Spherical} & \textbf{NinaPro} & \textbf{FSD50K} & \textbf{Darcy Flow} & \textbf{PSICOV} & \textbf{Cosmic} & \textbf{ECG} & \textbf{Satellite} & \textbf{DeepSEA} \\
\midrule
Backbone & Swin & Swin & Swin & Swin & Swin &  RoBERTa & Swin &  RoBERTa &  RoBERTa &  RoBERTa \\
Batch size & 32 & 32 & 32 & 32 & 4 & 1 & 4 & 4 & 16 & 16 \\
Epoch & 60 & 60 & 60 & 100 & 100 & 10 & 60 & 15 & 60 & 13 \\
Accumulation & 32 & 4 & 1 & 1 & 1 & 32 & 1 & 16 & 4 & 1 \\
Optimizer & SGD & AdamW & Adam & Adam & AdamW & Adam & AdamW & SGD & AdamW & Adam \\
Learning rate & 1.00E-04 & 1.00E-04 & 1.00E-04 & 1.00E-04 & 1.00E-03 & 5.00E-06 & 1.00E-03 & 1.00E-06 & 3.00E-05 & 1.00E-05 \\
Weight decay & 1.00E-03 & 1.00E-01 & 1.00E-05 & 5.00E-05 & 5.00E-03 & 1.00E-05 & 0.00E+00 & 1.00E-01 & 3.00E-06 & 0.00E+00 \\
Label discretization &- & - & - & - & 10 & 10 & 10& 10 &- & -  \\
\midrule
$\omega$ & 0.3 & 0.3 & 0.3 & 0.3 & 0.3 & 0.5 & 0.3 & 0.5 & 0.5 & 0.5 \\
FA epoch & 60 & 60 & 60 & 60 & 60 & 60 & 60 & 60 & 60 & 60 \\
FLD epoch & 4 & 4 & 4 & 4 & 4 & 4 & 4 & 2 & 4 & 2 \\
\bottomrule
\end{tabular}
}
\end{table*}

\begin{table*}[htbp]
\centering
\caption{ Hyperparameter configurations used for the 8 tasks on PDEBench~\citep{Takamoto2022PDEBENCHAE}.~{\em FA epoch} and {\em FLD epoch} denote the number of training epochs used to minimize the feature alignment (\textbf{FA}) and feature-label distortion (\textbf{FLD}).~The optimizers used in our experiments include SGD, Adam~\citep{kingma2017adammethodstochasticoptimization} and AdamW~\citep{loshchilov2017fixing}.}\vspace{2mm}
\label{pde_hyper}
\resizebox{\textwidth}{!}{
\begin{tabular}{|l||l|l|l|l|l|l|l|l|}
\toprule
\textbf{Hyperparameter} & \textbf{Advection} & \textbf{Burgers} & \textbf{RD1D} & \textbf{Diff-Sor} & \textbf{Navier-Stokes} & \textbf{Darcy-Flow} & \textbf{Shallow-Water} & \textbf{RD2D} \\
\midrule
Backbone &  RoBERTa &  RoBERTa &  RoBERTa & Swin &  RoBERTa & Swin & Swin & Swin  \\
Batch size & 4 & 4 & 4 & 4 & 4 & 4 & 4 & 4 \\
Epoch & 200 & 200 & 200 & 200 & 200 & 100 & 200 & 200 \\
Accumulation & 1 & 1 & 1 & 1 & 1 & 1 & 1 & 1 \\
Optimizer & Adam & Adam & SGD & AdamW & AdamW & AdamW & AdamW & Adam \\
Learning rate & 1.00E-04 & 1.00E-05 & 1.00E-03 & 1.00E-04 & 1.00E-04 & 1.00E-04 & 1.00E-04 & 1.00E-04 \\
Weight decay & 1.00E-05 & 1.00E-05 & 1.00E-05 & 0 & 1.00E-03 & 1.00E-05 & 0 & 1.00E-03 \\
Label discretization & 10 & 10 & 10 & 10 & 10 & 10 & 10 & 10 \\
\midrule
$\omega$ & 0.5 & 0.5 & 0.5 & 0.3 & 0.5 & 0.3 & 0.3 & 0.3  \\
FA epoch & 60 & 60 & 60 & 60 & 60 & 60 & 60 & 60 \\ 
FLD epoch & 4 & 4 & 4 & 4 & 4 & 4 & 4 & 4 \\
\bottomrule
\end{tabular}
}
\end{table*}

\section{Complexity Analysis}
We provide the complexity analysis of our proposed method as follows:

\textbf{Stage 1: Learning Feature Map:} Computing the loss $L_{\textbf{FA}}(\phi)$ in Eq.~\eqref{eq:FA-loss} incurs $O(f_\phi*|D_\tau|+f_\theta*|D_s|+|W_{\delta}|)$ processing cost where $f_\phi$ and $f_\theta$ are the forward costs of the source and the target embedder, respectively; $O(|W_{\delta}|)$ is the cost of computing the Wasserstein distance $W_{\delta}$.~Therefore, learning the target embedder within $n_1$ epochs with loss $L_{\textbf{FA}}(\phi)$ requires $O(f_\theta*|D_s|+n_1 (f_\phi*|D_\tau|+|W_{\delta}|+b_\phi))$ where $b_\phi$ is the cost of backpropagation of the target embedder.~Likewise, updating the target embedder within $n_2$ epochs with loss $L_{\textbf{FLD}}(\phi)$ in Eq.~\eqref{eq:FLD-loss} incurs $O(n_2(f_{p_s}|D_\tau|+t_{FLD}+b_{p_s,\phi}))$ where $f_{p_s}$ is the cost of attaining $z \sim p_s(z|\phi(\textbf{x}'))$, $t_{FLD}$ is the cost of computing the conditional entropy in Eq.~\eqref{eq:FLD-loss}, and $b_{p_s,\phi}$ is the cost of backpropagating through the frozen source predictor and target embedder.

\textbf{Stage 2: Learning Target Predictor:} Updating the target predictor with the loss $L_{\textbf{TF}}$ in $n_0$ epochs requires $O(n_0(f_{p_\tau}|D_\tau| + |D_\tau| +b_{p_\tau}))$ where $f_{p_\tau}$ and $b_{p_\tau}$ are forward and backpropagation cost of the target predictor. 

Therefore, the total computational cost of our proposed algorithm, RECRAFT, is:
\begin{eqnarray}
O(f_\theta*|D_s|+n_1 (f_\phi*|D_\tau|+|W_{\delta}|+b_\phi)) \ +\  O(n_2(f_{p_s}|D_\tau|+t_{FLD}+b_{p_s,\phi}))\ +\  O(n_0(f_{p_\tau}|D_\tau| + |D_\tau| +b_{p_\tau})) \nonumber
\end{eqnarray}

\begin{figure}[t]
    \vspace{.3in}
    \centering
    \includegraphics[width=0.9\textwidth]{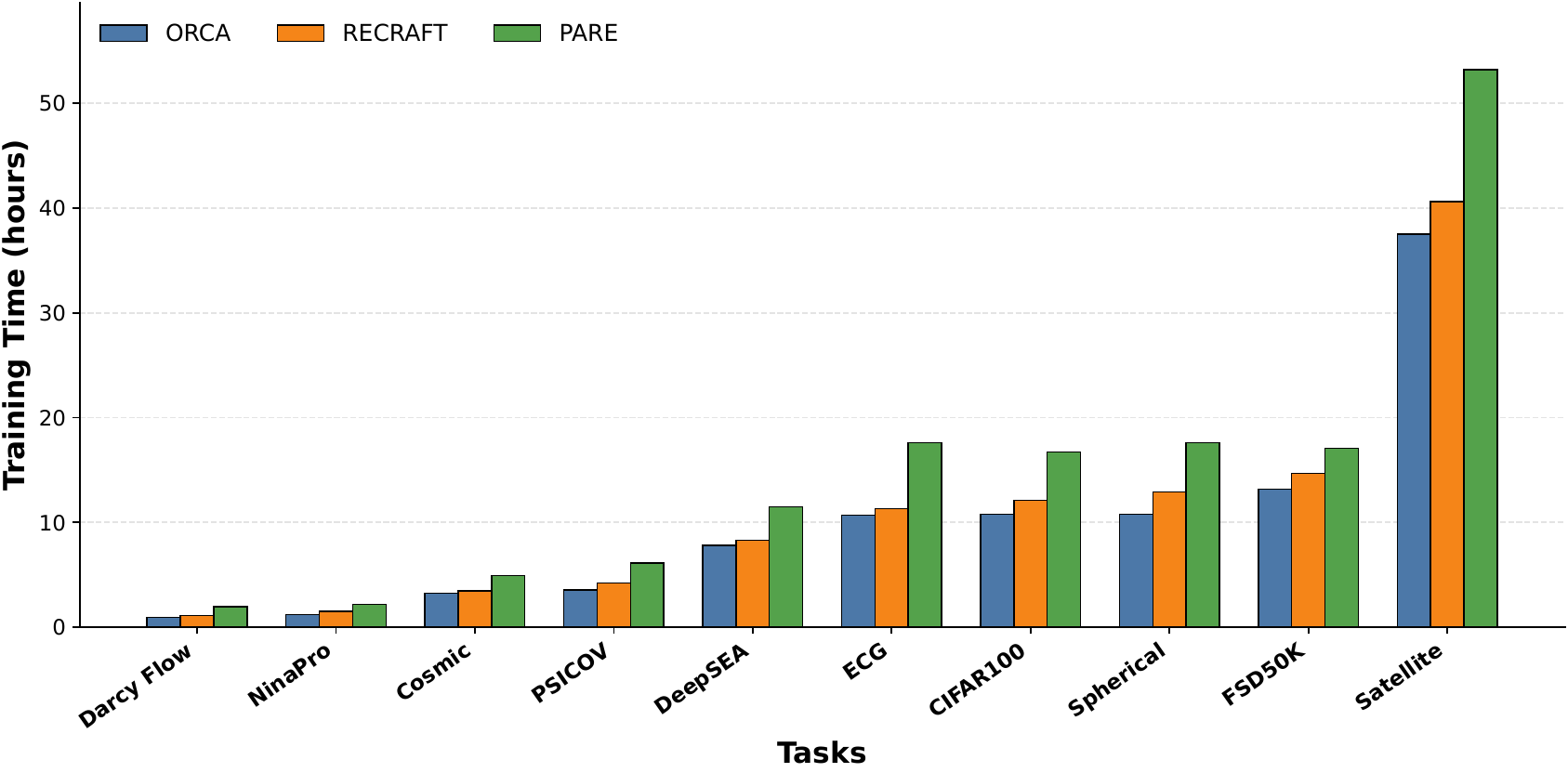}
    \caption{ Plots of training time comparison on NAS-Bench-360 across a variety of baselines.~RECRAFT achieves training times comparable to ORCA~\citep{shen2023orca}, while PARE~\citep{cai2024enhancingcrossmodalfinetuninggradually} incurs significantly higher training times across all tasks.}
    \label{tab:training_time}
\end{figure}

Compared to ORCA, RECRAFT introduces an additional computational cost of $O(n_2(f_{p_s}|D_\tau|+t_{FLD}+b_{p_s,\phi}))$ under identical settings for hyperparameters, number of training epochs, and Wasserstein distance $W_{\delta}$. However, this additional cost is empirically small, as demonstrated in Figure 5. Specifically, we compare the training efficiency (measured via training time) of RECRAFT against other state-of-the-art cross-modal fine-tuning baselines, including ORCA~\citep{shen2023orca} and PARE~\citep{cai2024enhancingcrossmodalfinetuninggradually}, using the NAS-Bench-360 benchmark~\citep{tu2022nasbench}. MoNA~\citep{ma2024learningmodalityknowledgealignment} is excluded from the comparison due to the unavailability of its implementation. As shown in Figure~\ref{tab:training_time}, RECRAFT achieves training times comparable to ORCA while delivering superior performance. In contrast, PARE incurs significantly higher training times across all evaluated tasks. Notably, RECRAFT outperforms PARE on 7 out of 10 tasks in NAS-Bench-360~\citep{tu2022nasbench}. These results suggest that RECRAFT offers the best trade-off between computational efficiency and fine-tuning performance.

\section{Benchmark Information}
\label{sec:benmark_info}
NAS-Bench-360~\citep{tu2022nasbench} encompasses ten tasks categorized into three groups: 2D classification, 2D dense prediction, and 1D classification. These tasks span specialized modalities, including protein sequences (PSICOV), PDE solving (Darcy-Flow), audio processing (FSD50K), genetic data analysis (DeepSEA), and electrocardiogram signals (ECG), among others. Table~\ref{tab:Nas_details} presents the details of each task, with a more comprehensive description available in the original paper~\citep{tu2022nasbench}.
\begin{table*}[htbp]
\centering
\small
\caption{\small Data statistics of the $8$ tasks in PDEBench~\citep{Takamoto2022PDEBENCHAE}.}
\label{tab:pde_infor}

\resizebox{\textwidth}{!}{
\begin{tabular}{|l|llllllll|}
\toprule
 \textbf{Task} & \textbf{Advection} & \textbf{Burgers} & \textbf{Diffusion-Reaction} & \textbf{Diffusion-Sorption} & \textbf{Navier-Stokes} & \textbf{Darcy-Flow} & \textbf{Shallow-Water} & \textbf{Diffusion-Reaction } \\
\midrule
\text{Input Shape} & 1D & 1D & 1D & 1D & 1D & 2D & 2D & 2D \\
\text{Output Type} & Dense & Dense & Dense & Dense & Dense & Dense & Dense & Dense \\
\text{Resolution} & 1024 & 1024 & 1024 & 1024 & 1024 & 128 $\times$ 128 & 128 $\times$ 128 & 128 $\times$ 128 \\
\text{Parameters} & $\beta = 0.4$ & $\nu = 1.0$ & $\nu = 0.5$, $\rho = 1.0$ & -- & $\eta = 1.0$, $\zeta = 1.0$ & $\beta = 0.1$ & -- & -- \\
\text{Loss} & nRMSE & nRMSE & nRMSE & nRMSE & nRMSE & nRMSE & nRMSE & nRMSE \\
\bottomrule
\end{tabular}
}
\end{table*}

\begin{table*}[t]
\centering
\caption{ \small Data statistics of the $10$ tasks in NAS-Bench-360~\citep{tu2022nasbench}.}
\label{tab:Nas_details}
\resizebox{\textwidth}{!}{
\begin{tabular}{|l|llllllllll|}
\toprule
& \textbf{CIFAR100} & \textbf{Spherical} & \textbf{NinaPro} & \textbf{FSD50K} & \textbf{Darcy Flow} & \textbf{PSICOV} & \textbf{Cosmic} & \textbf{ECG} & \textbf{Satellite} & \textbf{DeepSEA} \\
\midrule
\textbf{\# Training data} & 60K & 60K & 43956 & 51K & 1.1K & 3606 & 5250 & 330K & 1M & 250K \\
Input shape & 2D & 2D & 2D & 2D & 2D &  1D & 2D &  1D &  1D &  1D \\
\midrule
Output type & Point & Point & Point & Point & Dense &  Dense & Dense &  Point &  Point &  Point \\
\textbf{\# Classes} & 100 & 100 & 18 & 200 & -- & -- & -- & 4 & 24 & 36 \\
Loss & CE & CE & LpLoss & MSELoss & BCE & FocalLoss & BCE & CE & CE & BCE \\
\midrule
Expert Network & DenseNet-BC & S2CN & Attention Model & VGG & FNODE & EPCON & deepCR-mask & ResNet-1D & ROCKET & DeepSEA \\
\bottomrule
\end{tabular}
}
\end{table*}
PDEBench~\citep{Takamoto2022PDEBENCHAE} comprises multiple scientific datasets with simulated data from a wide variety of partial differential equations (PDEs) in physics. These include the 1D Advection equation, modeling linear advection with a constant speed parameter $\beta$; the 1D Burgers’ equation, capturing non-linear fluid dynamics with a constant diffusion coefficient $\nu$; the 1D Diffusion-Reaction equation, combining diffusion and a source term governed by parameters $\nu$ and $\rho$; and the 1D Diffusion-Sorption equation, representing diffusion retarded by sorption, applicable to real-world scenarios. Additionally, the 1D Navier-Stokes equation describes compressible fluid dynamics with shear and bulk viscosities $\eta$ and $\zeta$. In two dimensions, the Darcy-Flow equation models steady-state flow over a unit square, scaled by a constant force term $\beta$; the Shallow-Water equations, derived from Navier-Stokes, address free-surface flow problems; and the 2D Diffusion-Reaction equation extends its 1D counterpart with two non-linearly coupled variables, presenting a complex challenge with significant real-world applications. Table~\ref{tab:pde_infor} presents the details of each task, with a more comprehensive description available in the original paper~\citep{Takamoto2022PDEBENCHAE}.
\section{Boarder Impacts}
\label{sec:boarder_impacts}
Our research provides a new analytical lens that opens several promising research directions:

\noindent {\bf (i) Knowledge Distillation (KD).}~The bound naturally extends to KD by decomposing the teacher–student gap into feature alignment (FA) and softened-label distribution mismatch (FLD). This perspective suggests a new class of KD objectives that jointly minimize both components, potentially yielding more effective and robust distillation compared to standard temperature-scaled KL or feature-mimicry losses.

{\bf (ii) Retrieval-Augmented Generation (RAG) and Multimodal Retrieval.}~The same decomposition can guide the alignment of retrieved documents or heterogeneous data modalities (e.g., audio, video, or scientific figures) within a shared representation space.~This offers a theoretically grounded alternative to CLIP-style contrastive learning, with the potential for improved computational efficiency and robustness under label shift.

{\bf (iii) Scaling to Foundation Models and LLMs.}~Our framework highlights that explicitly accounting for FLD, beyond standard feature alignment, is critical for effective transfer. This insight provides a principled foundation for efficient cross-modal fine-tuning in large-scale models. Preliminary results on pre-trained Swin Transformer (90M parameters) and RoBERTa (88M+ parameters) demonstrate the feasibility of this approach and suggest a clear path toward scaling to modern foundation models.


\end{document}